\documentclass[journal]{IEEEtran}

\usepackage{moreverb,url}
\usepackage[colorlinks=false,bookmarksopen,bookmarksnumbered,hidelinks]{hyperref}
\usepackage{cite}
\usepackage{amsmath, amssymb, amsfonts}
\usepackage{mathptmx} 
\usepackage{array}
\usepackage{times} 
\usepackage[export]{adjustbox}
\usepackage{multirow}
\usepackage[flushleft]{threeparttable}
\usepackage{pdfcomment}
\usepackage{dblfloatfix}
\usepackage{booktabs}
\usepackage{siunitx}
\usepackage{tikz}

\usepackage{enumitem}

\let\oldcite\cite
\renewcommand{\cite}[1]{\mbox{\oldcite{#1}}}

\ifCLASSOPTIONcompsoc
    \usepackage[caption=false,font=normalsize,labelfont=sf,textfont=sf]{subfig}
\else
    \usepackage[caption=false,font=footnotesize]{subfig}
\fi

\ifCLASSINFOpdf
    \DeclareGraphicsExtensions{.pdf,.jpeg,.jpg,.png}
\fi

\usetikzlibrary{shapes, arrows}
\tikzstyle{block} = [draw, rectangle,
    minimum height=15mm, minimum width=15mm, align=center]
\tikzstyle{sum} = [draw, circle, minimum size=6mm]
\tikzstyle{input} = [coordinate]
\tikzstyle{output} = [coordinate]

\begin{document}

\title{A Computational Multi-Criteria Optimization Approach to Controller Design for\\ Physical Human-Robot Interaction}

\author{Yusuf~Aydin,~\IEEEmembership{Member,~IEEE,}\ %
        Ozan~Tokatli,~\IEEEmembership{Member,~IEEE,}\\%
        Volkan~Patoglu,~\IEEEmembership{Member,~IEEE,}\ %
        and~Cagatay~Basdogan,~\IEEEmembership{Member,~IEEE}
        \thanks{Y. Aydin and C. Basdogan are with the College of Engineering, Koc University, Istanbul, 34450 Turkey. E-mail: yaydin@ku.edu.tr, cbasdogan@ku.edu.tr.}
        \thanks{O. Tokatli is with RACE UKAEA, Abingdon, OX14 3DB, UK. E-mail: ozan.tokatli@ukaea.uk.}
        \thanks{V. Patoglu is with Faculty of Engineering and Natural Sciences, Sabanci University, Istanbul, 34956, Turkey. E-mail: vpatoglu@sabanciuniv.edu.}
        \thanks{Manuscript received October 31, 2019; revised March 22, 2020.}%
}

\markboth{IEEE Transactions on Robotics,~Vol.~XX, No.~X, XX~2020}%
{Aydin \MakeLowercase{\textit{et al.}}: A Computational Multi-Criteria Optimization Approach to Controller Design for Physical Human-Robot Interaction}

\maketitle

\begin{abstract}
Physical human-robot interaction (pHRI) integrates the benefits of human operator and a collaborative robot in tasks involving physical interaction, with the aim of increasing the task performance. However, the design of interaction controllers that achieve safe and transparent operations is challenging, mainly due to the contradicting nature of these objectives. Knowing that attaining perfect transparency is practically unachievable, controllers that allow better compromise between these objectives are desirable. In this paper, we propose a multi-criteria optimization framework, which jointly optimizes the stability robustness and transparency of a closed-loop pHRI system for a given interaction controller. In particular, we propose a Pareto optimization framework that allows the designer to make informed decisions by thoroughly studying the trade-off between stability robustness and transparency. The proposed framework involves a search over the discretized controller parameter space to compute the Pareto front curve and a selection of controller parameters that yield maximum attainable transparency and stability robustness by studying this trade-off curve. The proposed framework not only leads to the design of an optimal controller, but also enables a fair comparison among different interaction controllers. In order to demonstrate the practical use of the proposed approach, integer and fractional order admittance controllers are studied as a case study and compared both analytically and experimentally. The experimental results validate the proposed design framework and show that the achievable transparency under fractional order admittance controller is higher than that of integer order one, when both controllers are designed to ensure the same level of stability robustness.

\end{abstract}


\IEEEpeerreviewmaketitle

\section{Introduction}
\label{Sec:Introduction}

\IEEEPARstart{R}{obots} are superior to humans at tasks that require
precision, strength, and repetition, while the problem solving skills and
adaptability of humans are unmatched thanks to their cognitive abilities.
Physical human-robot interaction (pHRI) integrates the benefits of human and
robot in tasks that involve physical interaction~\cite{Ikeura1994}. From
assembly tasks to furniture relocation tasks in home/office setting~\cite{Mortl2012}, from
industrial applications~\cite{Wojtara2009}, surgery~\cite{Tavakoli2006},
rehabilitation~\cite{Pehlivan2016} to mission critical tasks such as
manipulation in hazardous environments~\cite{Reintsema2007}, collaboration of
humans and robots brings high performance solutions to complex problems~(please refer to
extensive reviews in~\cite{Goodrich2007,DeSantis2008,Ajoudani2018,MarciaReview}).
In such scenarios, the stability of the coupled human-robot dyad  must be ensured and the
controller should be sufficiently robust to the changes in both human and
environment dynamics for safe operation, while the dyad targets to achieve a
high task performance, which typically requires the robot to be transparent to the
human operator. However, the contradicting nature of robust stability and high
transparency requirements creates challenges during the design of interaction controllers.
In this study, we propose a computational multi-criteria optimization approach to design interaction controllers for pHRI to address the trade-off between stability robustness and transparency.

\subsection{Related Work}

The controller design for pHRI systems must pay utmost attention to the coupled
stability of the human-in-the-loop system. However, due to the complexity introduced by the presence of human operator and possible
contact interactions with uncertain environments, stability characteristics of
pHRI systems cannot be analysed easily.

Different approaches have been used to study coupled stability of interaction. Model based approaches~\mbox{\cite{Tsumugiwa2004,Duchaine2008,Gallagher2014,Dimeas2016,Aydin2018,FatihEmre2020}} rely on a simplified model of human dynamics and environment that capture most essential characteristics of these systems in terms of stability. In particular, given that the stiffness of the human arm and the environment have a direct effect on coupled stability~\mbox{\cite{Tsumugiwa2004,Dimeas2016,Aydin2018}}, most of these models focus on capturing this dominant aspect. In the absence of human and environment models, the coupled stability of pHRI systems can be investigated using the frequency domain passivity framework~\mbox{\cite{Colgate1988,Colgate1997,hulinStabPassBounds}}. In this approach, the human operator and the environment are assumed to act as passive elements that do not inject energy to the closed-loop system; hence, do not tend to destabilize the closed-loop system. While frequency domain passivity can guarantee the stability of the closed-loop system for a broad range of human/environment models, the resulting controllers perform conservatively~\mbox{\cite{Colgate1997,hulinStabPassBounds,hannafordRyu,ryuTeleop}}, leading to a less transparent performance. Time domain passivity framework~\mbox{\cite{hannafordRyu}} aims to relax the conservativeness of the frequency domain passivity framework by continually estimating the exchanged energy between robot and human and/or environment through sensor measurements and dissipating any excess energy as needed.

Satisfying passivity throughout the interaction ensures that the robot behaviour is never active (i.e. energy is not generated), so that stability can be guaranteed. However, non-passive systems are not necessarily unstable~\cite{Buerger2001}. This is due to the fact that passivity only considers the phase bounds on the uncertainty and assumes that the magnitude of this uncertainty can be arbitrarily large.
In practice, many systems are not exposed to such large changes or uncertainties. The conservativeness of the passivity framework can be relaxed by taking advantage of partial knowledge (e.g., magnitude bounds) on the models of human and/or environment. For instance, even though the dynamics of human arm changes over time, human arm impedance is known to vary in a relatively limited range~\cite{Dolan1993,Tsuji1995}. Along these lines, Buerger and Hogan~\cite{Buerger2007} proposed a complementary stability approach to design
interaction controllers that can  maintain robust stability for bounded ranges of
impedances and coupled stability is ensured without
the need for passivity. Similarly, in~\cite{Haddadi2010}, it is
shown that passivity constraints can be relaxed when the bounds on environmental/human
impedances are known.

The trade-off between robust stability and transparency brings a challenge in pHRI tasks that require high transparency~\cite{Lawrence}. This trade-off, as well
as the factors affecting the transparency, have been investigated in
detail~\cite{zaad2,griffiths,hulin,Willaert2014,abbott2007,hirche,Peer}. As an indicator of the achievable transparency, Z-width, the bounds of the dynamic range of achievable
impedances, was proposed by Colgate and Brown~\cite{colgateZwidth} and
methods to improve the Z-width were studied in~\cite{weirZwidth,chawdaVelEst}. Yet, it is well
known that attaining perfect transparency is practically
infeasible~\cite{Lawrence}. Hence, while keeping coupled stability intact, a controller
allowing for an optimal compromise between transparency and robustness is
desirable for pHRI studies~\cite{Peer}.

Recently, we proposed the use of fractional order admittance controllers (FOAC) for pHRI, compared their performance with conventional integer order
admittance controllers (IOAC) and provided evidence that FOAC can offer better  stability robustness
while displaying higher transparency than IOAC~\cite{yusufWHC}.
In particular, we introduced the concept of  impedance matching to enable comparisons between FOAC and
IOAC  at a given frequency and analytically showed that both stability
robustness and transparency performance can be improved under FOAC~\cite{Aydin2018}.

Earlier studies have already shown that a compromise between
stability robustness and transparency is possible by manipulating the parameters
of an interaction controller for pHRI~\cite{Aydin2018,Buerger2007,yusufWHC}.
Nevertheless, analytical methods have proven to be prohibitive for the design of an optimal
interaction controller for pHRI, due to the non-trivial interaction between the control parameters and the dynamics of the coupled system. On the other hand, computational approaches have been quite
promising. Computational optimization  approaches have been proposed in~\mbox{\cite{Buerger2007,Labrecque2015}} to design controllers with improved performance. While stability and transparency have been considered in these studies, both studies address single objective optimization problems. In particular, in~\mbox{\cite{Buerger2007}} stability is imposed as a constraint, while in~\mbox{\cite{Labrecque2015}} multiple criteria are aggregated into a single cost function through predetermined weights.

\subsection{Contributions} \label{SubSec:Contributions}

In this study, we propose a computational multi-criteria optimization approach
to design linear time-invariant (LTI) interaction (i.e., admittance and impedance) controllers for pHRI tasks. To our knowledge, no such multi-criteria optimization
framework has been proposed to \emph{simultaneously} optimize interaction
controllers for multiple objectives, such as stability robustness and transparency.

Since the dynamics of interaction among human, robot and the environment are highly complex, and rendering the development of an analytical approach is prohibitive, a computational approach is proposed. Earlier computational studies for optimizing the human-robot interaction have focused on
determination of joint manipulation trajectory~\cite{Norouzzadeh2012,Losey2018}, improvement of task
performance~\cite{Passenberg2011,Rahman2016,Hulin2017}, collaborative task
planning and scheduling~\cite{Gombolay2017,Pearce2018}, but not on the trade-off
between robust stability and transparency.

Buerger and Hogan~\cite{Buerger2007}, and Labrecque and
Gosselin~\cite{Labrecque2015} have proposed \emph{single criterion optimization} of interaction
controllers. In these approaches, the original multi-objective problem is formulated as a single
criterion optimization problem, either by considering stability (robustness) as a constraint and/or by scalarizing the cost functions using predetermined weights to define a single aggregate objective function. The drawback of these approaches is that the preferences between objectives need to be assigned a priori, before having a complete knowledge on the trade-offs involved.

On the other hand, the proposed  Pareto optimization approach computes all possible \emph{non-dominated} solutions that form the Pareto front curve and allows the designer to make informed decisions on \emph{optimal} interaction controllers.  Unlike one-shot scalarization-based optimization methods, Pareto methods fully characterize the trade-off among objectives by providing all optimal solutions for all possible preferences of the designer. This allows an optimal solution to be selected after the trade-off is thoroughly studied, possibly by considering new criteria that have not been considered during optimization. Furthermore, Pareto methods allow the designer to choose alternative optimal solutions under different conditions.

In addition to multi-criteria optimal design of interaction controllers, the use of Pareto  optimization approach enables \emph{fair comparisons} among different controller structures, as comparison can be performed based on the best possible performance of controllers for all possible user preferences. We demonstrate such a comparison as a case study that compares FOAC and IOAC and by experimentally verifying that achievable transparency under FOAC is higher than that of IOAC, when same level of stability robustness is required.

\smallskip
The rest of this paper is organized as follows:
Section~\ref{Sec:ComputationalDesignFramework} introduces our computational interaction controller design approach. Section~\ref{Sec:CaseStudy} presents running examples for which the
interaction control architectures, and transparency and stability robustness
metrics are introduced. Pareto optimization is presented in
Section~\ref{SubSec:MulticriteriaOptimization}. Human subject experiments
conducted to assess the transparency under optimal FOAC and
IOAC are reported in Section~\ref{Sec:Experiments}. Discussions and conclusions are provided in Section~\ref{Sec:Conclusion}.

\section{Multi-Criteria Design Optimization Approach}
\label{Sec:ComputationalDesignFramework}

\subsection{Designing Controllers for Multiple Objectives}
\label{SubSec:ControllerDesign}

Stability robustness and transparency performance analyses of interaction
controllers are challenging, as the standard tools used for servo control cannot
be applied to controller design. Two fundamental differences exist for interaction control systems
as noted in~\cite{Buerger2007}:
(i) the closed-loop stability and performance cannot be predicted and
characterized using the open-loop transfer function alone, and (ii) the
controller parameters do not directly affect the stability and performance of
the coupled system. Consequently, there exists no straightforward way to
investigate the stability and performance of such systems analytically, while
computational approaches have been shown to provide promising
results~\cite{hulinStabPassBounds,Buerger2007,Labrecque2018}.

Buerger and Hogan~\cite{Buerger2007} used a computational loop-shaping approach
to optimize an interaction controller for maximum performance while satisfying a stability
constraint. In their approach, first, the stability of the closed-loop system is evaluated in the controller parameter space that covers all possible combinations of controller parameters and then, the controller parameters that yield the maximum performance among the stable solutions are selected.

We advocate Pareto methods for multi-criteria optimization of
interaction controllers. Given a multi-criteria optimization problem, all
non-dominated solutions constitute the Pareto front~\cite{papalambros_wilde_2000,Marler2010,unalPareto}. Every Pareto optimization approach consists of three main steps: i) selection and evaluation of cost functions, ii) computation of the Pareto front, and iii) selection of an optimal solution among all non-dominated solutions.
Below we provide an outline of the proposed approach.

\vspace{-.5\baselineskip}
\subsection{Objective Functions for Interaction Controllers}
\label{SubSec:Objectives}

Depending on the task, the performance of a pHRI application can be assessed using various criteria. In this study, we focus on the two fundamental ones which are known to possess an inherent trade-off: stability robustness and transparency.

The first objective function is chosen as the stability robustness of the closed-loop
system. The stability is an integral part of robotic applications, especially in
pHRI domain, where it implies inherent safety of the operator. For pHRI applications, a
degree of robustness in the stability of the closed-loop system needs to be guaranteed, since
elements of the closed-loop transfer function, e.g. human and environment, are subject to change during the task execution. Under these circumstances, a controller that can provide closed-loop stability over a desired
range of parameter variations is desirable.

The second objective is chosen as the transparency of the pHRI task. The impedance
that a human feels during interactions with an environment has critical
importance in many pHRI tasks. For instance, in a robotic assisted surgery, the
control architecture for the interaction is expected to ensure that the impedance of the
soft tissue is reflected to the surgeon so that she/he can have a realistic understanding of
the operation, without being shadowed by the dynamics of the robot. Alternatively, in a
different task, an operator may wish to move the robot in free space, where she/he
does not desire to feel any parasitic dynamics due to the robot.

\subsection{Pareto Optimization}
\label{SubSec:MulticriteriaOptimization}

Assuming an LTI model of the robot is available and an LTI controller structure to
regulate the interaction between the robot and the human is selected, the
following steps are proposed to optimize the controller parameters for a robust
and transparent design.

\begin{enumerate}[leftmargin=1.2cm,label={Step \arabic*)}]
    \item A feasible range of values for each parameter of the interaction controller is chosen (and discretized). Feasible controller parameters are computed as design variables by considering these range of values for each parameter.

    \item A metric for transparency is defined  for the design variables in Step 1.

    \item A metric for stability robustness is defined for the design variables in Step 1.
    \item Using a Pareto optimization approach, the Pareto front curve is constructed by considering the transparency and robustness objectives.
\end{enumerate}

While many different approaches have been proposed in the literature to compute the Pareto front,  we utilize the weighted-sum approach with weight scanning~\cite{Marler2010} in this study, as this approach is easy to implement and present.

\subsection{Selection of Optimal Controller}
\label{SubSec:DesignSelection}

The resulting Pareto front represents all non-dominated solutions for the given design problem
and these solutions reflect optimal solutions for different preferences among the selected metrics. Once the Pareto front is computed, the designer is expected to study these solutions to get an insight of the underlying trade-offs and make an informed decision to finalize the controller design by selecting an optimal solution from the Pareto set. Constraints or new criteria that have not been considered during the optimization can be introduced at this stage. Pareto methods allow the designer to choose alternative optimal solutions under different conditions.

\smallskip

The Pareto optimization approach is rewarding as it not only leads to design of optimal interaction controllers but also enables fair comparisons among various interaction controllers, possibly with different underlying types or structures. Given that the Pareto front for each controller provides the best performance of that controller for all possible preferences of the designer, fair comparisons become possible by considering the Pareto front of each controller type/structure. Hence, for any given designer preference, the best possible performances of each controller is compared to the best possible performances of other controllers.

\section{Case Study: Interaction Controller Design}
\label{Sec:CaseStudy}

This section presents a case study that follows the Pareto optimization approach
introduced in Section~II. In particular, a family of admittance controllers are studied in terms of stability robustness and transparency, and optimal controller parameters are determined from their Pareto solutions.

\subsection{pHRI Tasks}
\label{SubSec:pHRITasks}

Three different pHRI scenarios are considered to demonstrate the proposed
computational optimization framework in designing an interaction controller.
\begin{enumerate}
    \item[S1] Only the human and robot physically interact. Ideally, human
        operator desires not to feel the robot dynamics while moving the robot in free
        space.
    \item[S2] The human operator continually interacts with a nominal environment
    using the robot. Note that the dynamics of the environment does not change during the execution of the pHRI task (in other words, an LTI model for the environment is assumed).
    \item[S3] The human operator interacts with an environment using the
        robot, while the dynamics of the environment changes.
\end{enumerate}

S1 involves no interaction with environment, so this scenario is considered as a baseline for S2 and S3. In S2, the contact with environment involves a linear spring. In S3, the dynamics of the environment is changed by considering two springs in parallel, where one of the springs is engaged only during a portion of the interaction. Along these lines,  S3 involves switching dynamics resulting in a nonlinear interaction.

\subsection{The Closed-Loop System}
\label{SubSec:ControlSystem}
\begin{figure}[b]
    \centering
    \resizebox{\columnwidth}{!}{\rotatebox{0}{%
        \begin{tikzpicture}[auto, node distance=20mm,>=latex']
            \coordinate [] (aux1) {};
            \node [block, right of=aux1] (admCtrl) {Admittance\\Controller\\{\large $Y(s)$}};
            \node [sum, right of=admCtrl, node distance=25mm] (sum1) {};
            \node [block, right of=sum1] (robotMotionCtrl) {Robot\\Motion\\Controller};
            \node [sum, right of=robotMotionCtrl] (sum2) {};
            \node [block, right of=sum2] (robotDynamics) {Robot\\Dynamics};
            \node [block, yshift=-10mm, below of=robotMotionCtrl] (filter) {Filter\\{\large $H(s)$}};
            \coordinate [right of=filter] (aux2) {};
            \node [block, yshift=-15mm, node distance=25mm, right of=robotDynamics] (env) {Environment\\{\large $Z_e(s)$}};
            \node [sum, below of=env, yshift=5mm] (sum3) {};
            \node [block, right of=env, node distance=25mm] (human) {Human\\{\large $Z_{h}(s)$}};
            \node [sum, above of=human, yshift=-5mm] (sum4) {};
            \node [input, right of=sum4, node distance=10mm, above] (vDes) {};
            \coordinate [xshift=5mm, left of=env] (aux3) {};

            \draw [->] (admCtrl) node [xshift=13mm, above] {\Large $v_\text{ref}$} -- node [pos=0.90, below] {\large+}(sum1);
            \draw [->] (sum1) -- (robotMotionCtrl);
            \draw [->] (robotMotionCtrl) node [xshift=15mm, above] {\Large+} -- (sum2);
            \draw [->] (sum2) -- (robotDynamics);
            \draw [-] (filter) -| (aux1);
            \draw [->] (aux1) -- node [above] {\Large $F_\text{int}$}(admCtrl);
            \draw [->] (aux2) -- (filter);
            \draw [->] (aux2) -- node [pos=0.90, right] {\Large-} (sum2);
            \draw [->] (env) node [yshift=-10mm, left] {\Large $F_e$} -- node [pos=0.90, right] {\Large-} (sum3);
            \draw [-]  (sum3) -- (aux2);
            \draw [->] (human) |-  node [pos=0.6, below] {\Large $F_h$} node [pos=0.92, below] {\Large+} (sum3);
            \draw [->] (robotDynamics) -| (env);
            \draw [->] (sum4) -- (human);
            \draw [->] (robotDynamics) node [xshift=45mm, above] {\Large-} node [xshift=20mm, above] {\Large $v$}-- (sum4);
            \draw [->] (vDes) node [xshift=-2mm, above] {\Large $v_\text{des}$} node [xshift=-5mm, below] {\Large+} -- (sum4);
            \draw [-]  (robotDynamics) -| (aux3);
            \draw [->] (aux3) -| node [pos=0.90, right] {\Large-} (sum1);

            \draw [dash dot] (14.5, -4) -- (14.5, 1);
            \draw [->] (15.1, -4) -- node [right, xshift=2mm] {\large $Z_\text{disp}(s)$} (14.6, -4);
            \draw [dash dot] (11.8, -4) -- (11.8, 1);
            \draw [->] (12.4, -4) -- node [right, xshift=2mm] {\large $\Delta Z(s)$} (11.9, -4);

            \draw [dotted] (3.75, -1.75) rectangle (11.65, 1.5) node [below, xshift=-30mm] {\Large $G(s)$};
        \end{tikzpicture}
    }}\vspace{-\baselineskip}
    \caption{Control architecture of the pHRI system\label{Fig:ControlArchitecture}}
\end{figure}
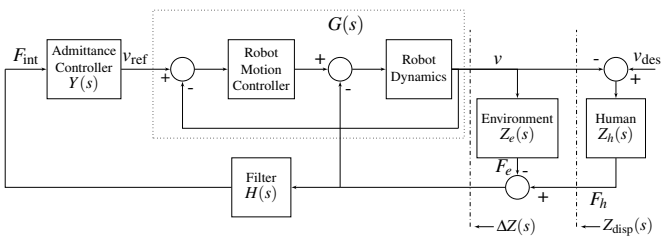

Figure~\ref{Fig:ControlArchitecture} depicts the admittance control architecture
used in our pHRI studies~\cite{yusufWHC, Aydin2018}, where the human, robot, and the
environment  physically interact with each other. The transfer function of
the closed-loop system is given by
\begin{align}
    T(s) = \frac{V(s)}{F_{\text{int}}(s)} = \frac{G(s)Y(s)}{1+G(s)Y(s)H(s)Z_{\text{eq}}(s)}
    \label{Eqn:closedLoopTF}
\end{align}

\noindent where $G(s)$ is the LTI model of robot, $V(s)$ is its measured
(actual) end-effector velocity, $Y(s)$ is the admittance controller, and $H(s)$
models a filter introduced to attenuate the noise in force measurements. Human and environment
are assumed to be coupled, and their equivalent impedance is set to
$Z_{\text{eq}}=Z_h+Z_e$, where $Z_h$ and $Z_e$ represent linearized human and
environment impedances, respectively. The equivalent impedance used in our study
is $Z_{\text{eq}}(s)=\frac{m_{\text{eq}}s^2+b_{\text{eq}}s+k_{\text{eq}}}{s}$,
where equivalent stiffness, damping, and mass elements can be defined as
$k_{\text{eq}}=k_h+k_e$, $b_{\text{eq}}=b_h+b_e$, and $m_{\text{eq}}=m_h+m_e$,
respectively.

FOAC used in this study has the following form
\begin{align}
    Y(s) = \frac{1}{Z_{\text{FOAC}}}=\frac{1}{m_Fs^\alpha+b_F}
    \label{Eqn:FOAdmittanceCtrl}
\end{align}

\noindent where $\alpha$ corresponds to the order of the fractional integrator
(i.e., the integration order), while $m_F$ and $b_F$ are the \emph{admittance
controller parameters}. In this study, the integration order is
kept in the range of $0 < \alpha \leq 1$. In the limit case when $\alpha=1$, the
FOAC becomes equivalent to IOAC; therefore, the admittance controller becomes $Y(s) =
1/Z_{\text{IOAC}}=1/(m_Fs+b_F)$.

As the unit of $m_F$ is kgs$^{\alpha-1}$, the physical interpretation of this parameter
changes depending on the integration order $\alpha$, whereas the unit of $b_F$
is always Ns/m. The effective mass and damping provided by FOAC are
$m_F\omega^{\alpha-1}\sin(\frac{\alpha\pi}{2})$ kg and $b_F+
m_F\omega^\alpha\cos(\frac{\alpha\pi}{2})$ Ns/m, respectively, where $\omega$ represents
the frequency. Thorough analyses of the effective impedance of the fractional order interaction controllers
can be found in~\cite{Aydin2018,yusufWHC,OzanIROS,OzanISRR}.
\begin{figure*}[ht]
    \centering
    \begin{minipage}{\textwidth}
        \centering
        \subfloat[$\alpha=1$\label{Fig:tMap1}]{%
            \resizebox{0.3\columnwidth}{!}{\rotatebox{0}{\includegraphics[trim=0cm 0.05cm 0.68cm 0.2cm, clip=true]{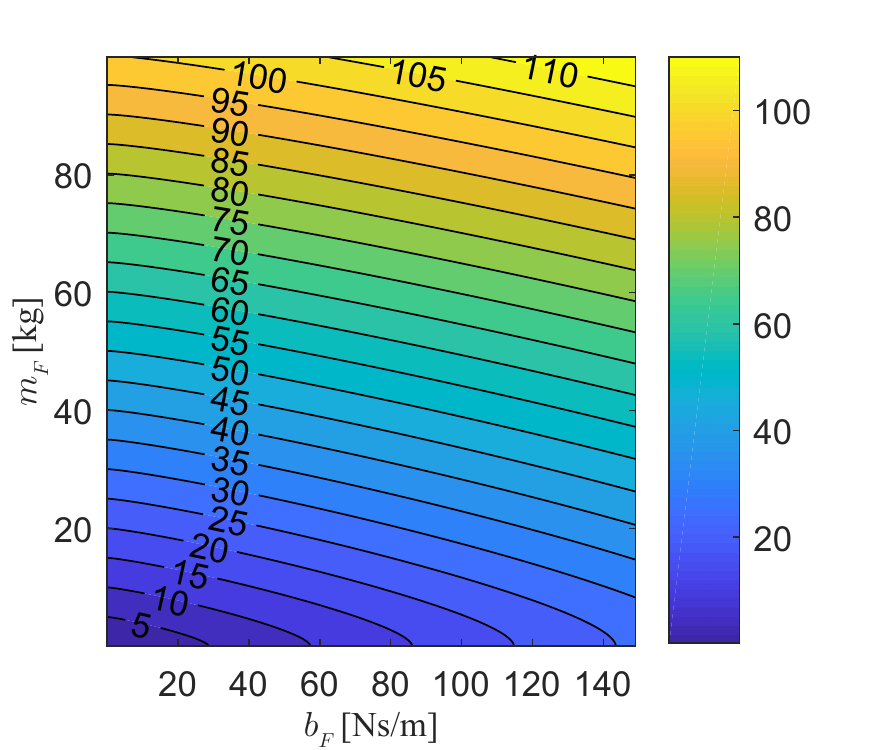}}}
        }
        \hfill
        \subfloat[$\alpha=0.7$\label{Fig:tMap7}]{%
            \resizebox{0.3\columnwidth}{!}{\rotatebox{0}{\includegraphics[trim=0cm 0.05cm 0.68cm 0.2cm, clip=true]{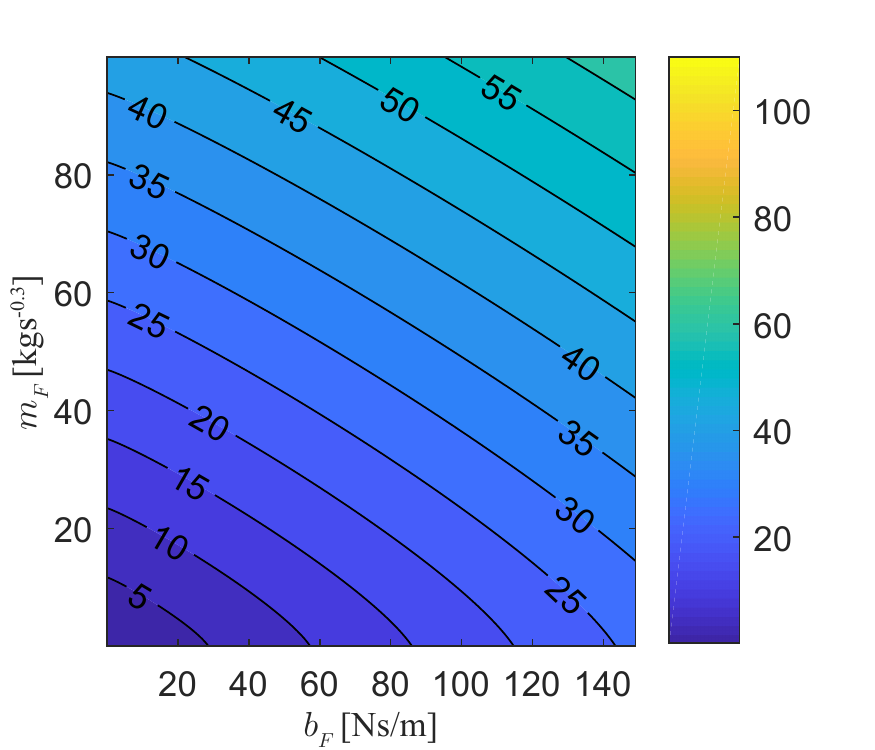}}}
        }
        \hfill
        \subfloat[$\alpha=0.4$\label{Fig:tMap4}]{%
            \resizebox{0.3\columnwidth}{!}{\rotatebox{0}{\includegraphics[trim=0cm 0.05cm 0.68cm 0.2cm, clip=true]{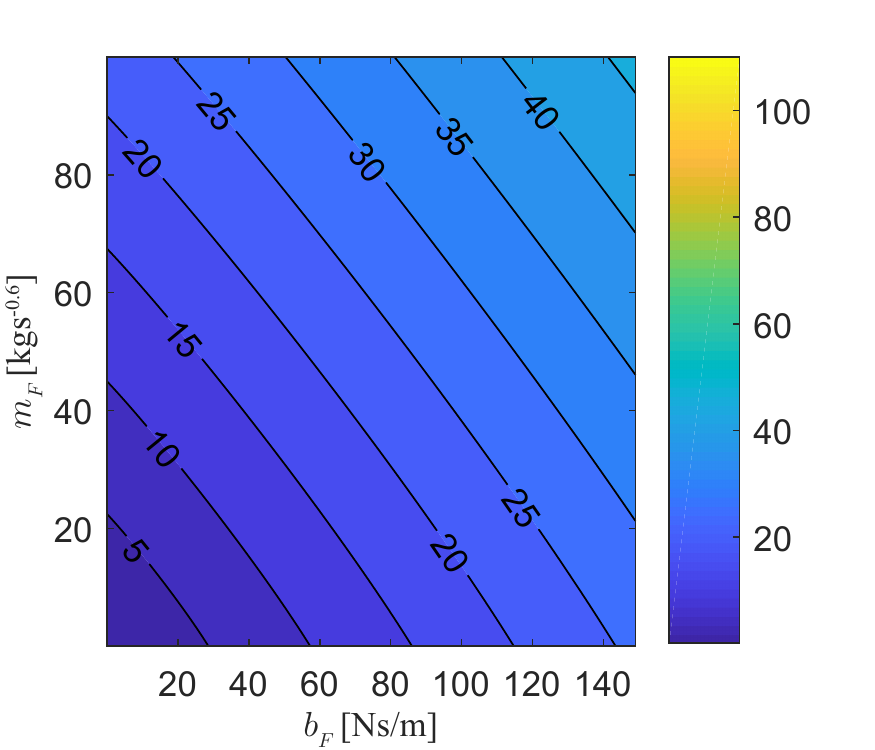}}}
        }
    \end{minipage}
    \vspace{-\baselineskip}
    \caption{Transparency maps illustrating the parasitic impedance of the controller in Figure~\ref{Fig:ControlArchitecture} under (a) IOAC $\alpha=1$, (b) FOAC $\alpha=0.7$, and (c) FOAC $\alpha=0.4$.}
    \label{Fig:transMaps} \vspace{-.5\baselineskip}
\end{figure*}

Knowing that controller parameters have different physical interpretations and the structure of an admittance controller is different for each integration order $\alpha$, a fair comparison among controllers of different orders is challenging. Therefore, being able to thoroughly investigate and rigorously compare the trade-off between stability robustness and transparency under various integration orders by using the optimization framework proposed in this study is a rewarding practice.

\subsection{Bounds on Equivalent Impedance ($Z_\text{eq}$)}
\label{SubSec:EquivalentImpedanceBounds}

In this section, we determine the bounds on equivalent impedance $Z_\text{eq}$ for each pHRI scenario.

For the first scenario (S1), only the bounds of the human arm impedance is
considered, since $Z_e(s) = 0$. Lower and upper bounds for
the human arm stiffness are taken as $k_h=0$ (when human is not in contact with
the end-effector of the robot) and $k_h=600$ N/m (based on~\cite{Dolan1993,Tsuji1995}), respectively. Moreover, the lower and upper bounds for the mass parameter are set to $m_h=0$ (when human releases
her/his contact) and $m_h=5$ kg, respectively. The range of damping for
the human arm $b_h$ is taken as 0 to 41 Ns/m, as proposed in~\cite{Dolan1993,Tsuji1995}.

For the second scenario (S2), the contact interaction with a spring-like
environment having a stiffness of $k_e$ is considered. Therefore, the damping, and
the mass of the equivalent impedance are kept the same as damping and mass of human
arm as in S1: $b_{\text{eq}}=b_h$ and $m_{\text{eq}}=m_h$. The environment is
assumed to have a nominal stiffness of $k_e=610$ N/m, thus the range for the equivalent
stiffness is set to $610\text{ N/m} \leq k_{\text{eq}} \leq 1210\text{ N/m}$.

Similarly, for the third scenario (S3), the environment stiffness varies between $610\text{ N/m} \leq k_e \leq 1010\text{ N/m}$, thus, the range for the equivalent stiffness is set to $610\text{
N/m} \leq k_{\text{eq}} \leq 1610\text{ N/m}$.

\subsection{Computation of the Pareto Front}
\label{SubSec:ObjectiveFunctions}
\bigskip
\textbf{Step 1} -- \emph{Feasible Controller Parameters}

For a given integration order $\alpha$, $m_F\in[0.2,100] \text{ kgs}^{\alpha-1}$ and $b_F \in[0.001,500]$ Ns/m are taken as the feasible range of the admittance controller parameters and these ranges are discretized with equally spaced increments of 0.1 $\text{ kgs}^{\alpha-1}$, and 1 Ns/m, respectively.
\bigskip
\bigskip

\textbf{Step 2} -- \emph{Transparency Cost Function}
\label{SubSec:Transparency}

The closed-loop impedance displayed to the human operator $Z_{\text{disp}}(s)$
according to Figure~\ref{Fig:ControlArchitecture} can be computed
as
\begin{align}
    Z_{\text{disp}}(s) = \frac{F_h(s)}{V(s)} = \frac{1+G(s)Y(s)H(s)Z_{e}(s)}{G(s)Y(s)H(s)}
    \label{Eqn:closedLoopImpedance}
\end{align}

The parasitic impedance $\Delta Z(s)$ is defined as the difference between the
desired impedance $Z_{\text{des}}(s)$ and the impedance reflected to human
$Z_{\text{disp}}(s)$:
\begin{align}
    \Delta Z(s) \triangleq Z_{\text{des}}(s)-Z_{\text{disp}}(s)
    \label{Eqn:diffImp}
\end{align}

If the parasitic impedance $\Delta Z(s)$ is small, then the
transparency of the overall system is high. In most pHRI applications, the
desired impedance is equal to the environment impedance,
$Z_{\text{des}}(s)=Z_e(s)$. In this condition, considering~\eqref{Eqn:closedLoopImpedance} and~\eqref{Eqn:diffImp}, the magnitude of
parasitic impedance in frequency domain can be determined as follows:
\begin{align}
    |\Delta Z(j\omega)| = 1/|G(j\omega)Y(j\omega)H(j\omega)|
    \label{Eqn:delZSimple}
\end{align}

\noindent Clearly, maximizing $|GYH|$ minimizes $|\Delta Z|$, which in turn
maximizes the transparency. Furthermore, in terms of the controller,
we can deduce that maximizing the magnitude of $Y(j\omega)$ is required to improve the transparency. According to $Y(j\omega)$, lower values of $m_F$ and $b_F$ result in higher transparency.
At low frequencies, the effect of $b_F$ is more dominant on the
parasitic impedance. Therefore, lower values of $b_F$ are more desirable for
higher transparency at low frequencies. Moreover, effects of $\alpha$ and $m_F$
become increasingly more dominant at higher frequencies.

While inspection of the magnitude of $Y(j\omega)$ provides insight about how
controller parameters affect the transparency, the parasitic impedance function should be studied
for more conclusive results, since the dynamics of robot
itself also contributes to the parasitic impedance. In particular, the fact that the contributions of controller and robot on parasitic impedance are complex numbers and change as functions of frequency calls for a
quantitative cost metric for the parasitic impedance. Such a metric was defined in~\cite{Buerger2007} and computed over a discrete range of frequencies. Adopting
this metric, the transparency cost function is defined as
\begin{align}
    C = \sum_{\omega_L}^{\omega_U}W(\omega)\log |\Delta Z(j\omega)|
    \label{Eqn:parasiticcost}
\end{align}
\noindent where $W(\omega)$ is a weighting function defined to adjust
contributions at each frequency, and $\omega_L$ and $\omega_U$ represent the lower and upper bounds of the frequency range of interest, respectively.

In this study, a logarithmically spaced frequency ranging from
\SIrange{0.01}{30}{\hertz} is chosen for the discretization. Since the frequency range of human voluntary
movements is around 2~Hz~\cite{Brooks1990}, achieving  higher transparency is more desirable at lower
frequencies. Along these lines, a fifth order Butterworth filter with a
cut-off frequency of 5~Hz is used as the weighting function $W(\omega)$ to boost the effect of
low frequency content on the parasitic impedance and reduce the
contribution of higher frequencies.  Using~\eqref{Eqn:parasiticcost}, parasitic impedance for each
set of controller parameters is evaluated, and the resulting transparency maps illustrating the parasitic impedance are constructed as in Figure~\ref{Fig:transMaps}. As expected, lowering
$m_F$, and $b_F$ reduce the parasitic impedance, leading to higher
transparency.

\medskip
\textbf{Step 3} -- \emph{Stability Robustness Cost Function}
\label{SubSec:RobustStability}
\begin{figure*}[t]
    \centering
    \begin{minipage}{\textwidth}
        \centering
        \subfloat[$\alpha=1$, $m_{\text{eq}}=0$ kg, \newline $b_{\text{eq}}=0$ Ns/m\label{Fig:rob1_1}]{%
            \resizebox{0.24\columnwidth}{!}{\rotatebox{0}{\includegraphics[trim=0cm 0.05cm 1.05cm 0.2cm, clip=true]{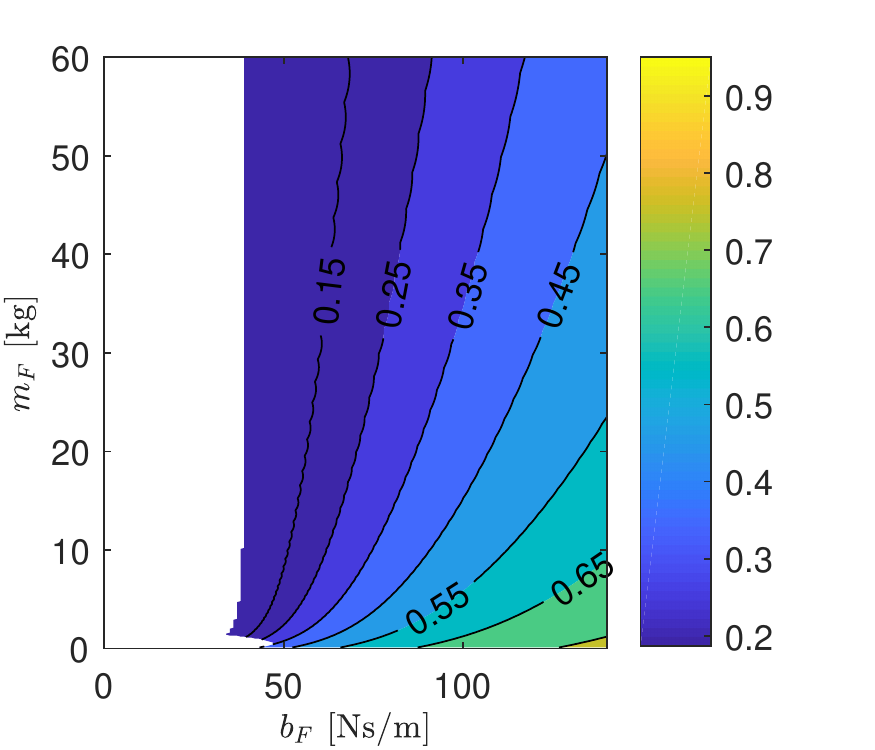}}}
        }
        \hfill
        \subfloat[$\alpha=1$, $m_{\text{eq}}=5$ kg, \newline $b_{\text{eq}}=0$ Ns/m\label{Fig:rob7_1}]{%
            \resizebox{0.24\columnwidth}{!}{\rotatebox{0}{\includegraphics[trim=0cm 0.05cm 1.05cm 0.2cm, clip=true]{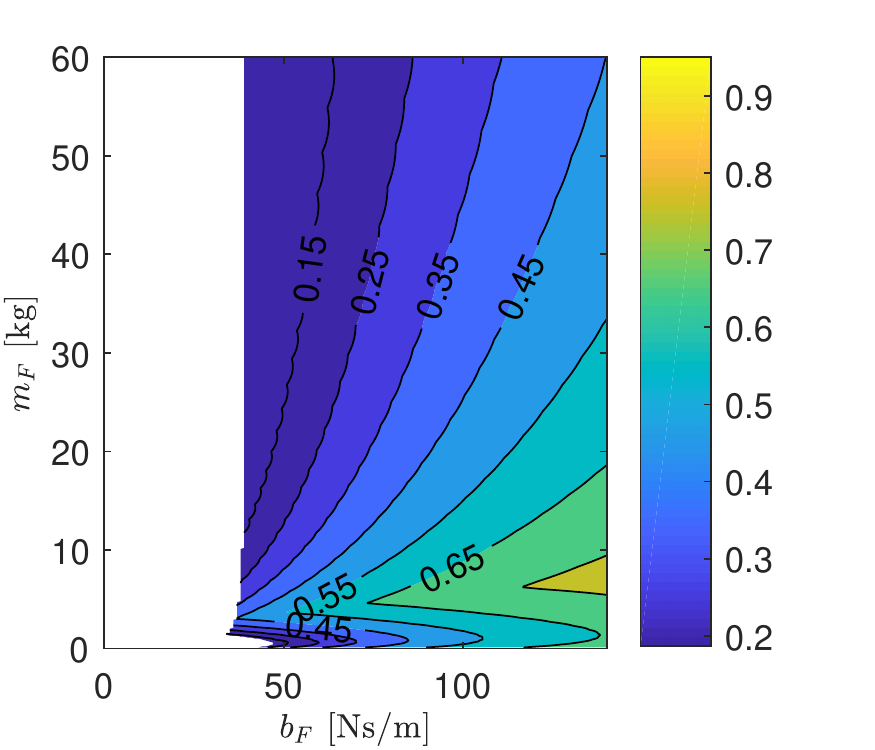}}}
        }
        \hfill
        \subfloat[$\alpha=1$, $m_{\text{eq}}=0$ kg, \newline $b_{\text{eq}}=41$ Ns/m\label{Fig:rob4_1}]{%
            \resizebox{0.24\columnwidth}{!}{\rotatebox{0}{\includegraphics[trim=0cm 0.05cm 1.05cm 0.2cm, clip=true]{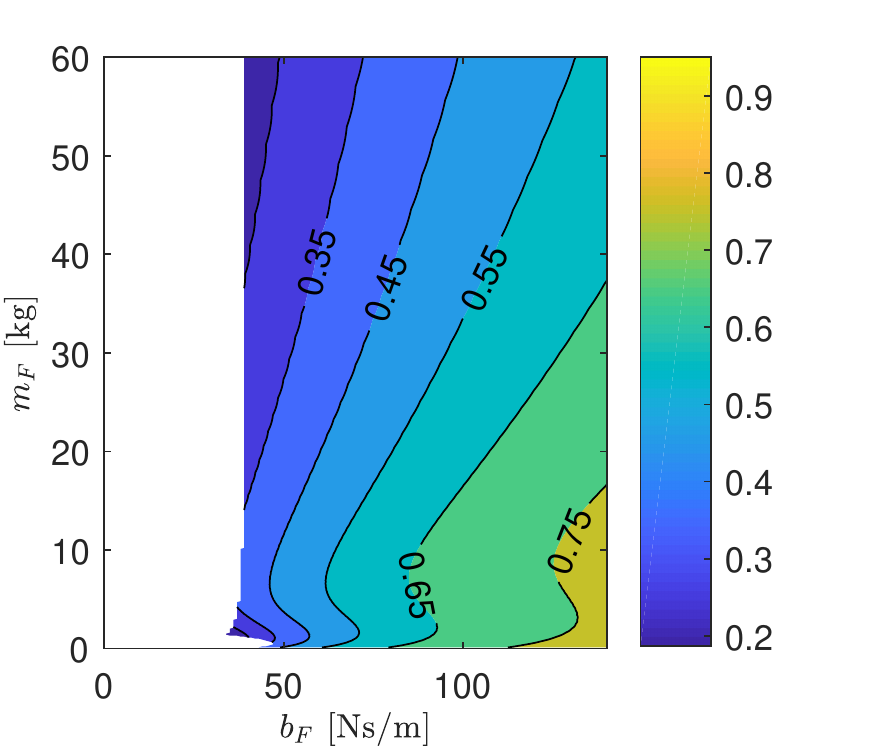}}}
        }
        \hfill
        \subfloat[$\alpha=1$, $m_{\text{eq}}=5$ kg, \newline $b_{\text{eq}}=41$ Ns/m\label{Fig:rob4_1}]{%
            \resizebox{0.24\columnwidth}{!}{\rotatebox{0}{\includegraphics[trim=0cm 0.05cm 1.05cm 0.2cm, clip=true]{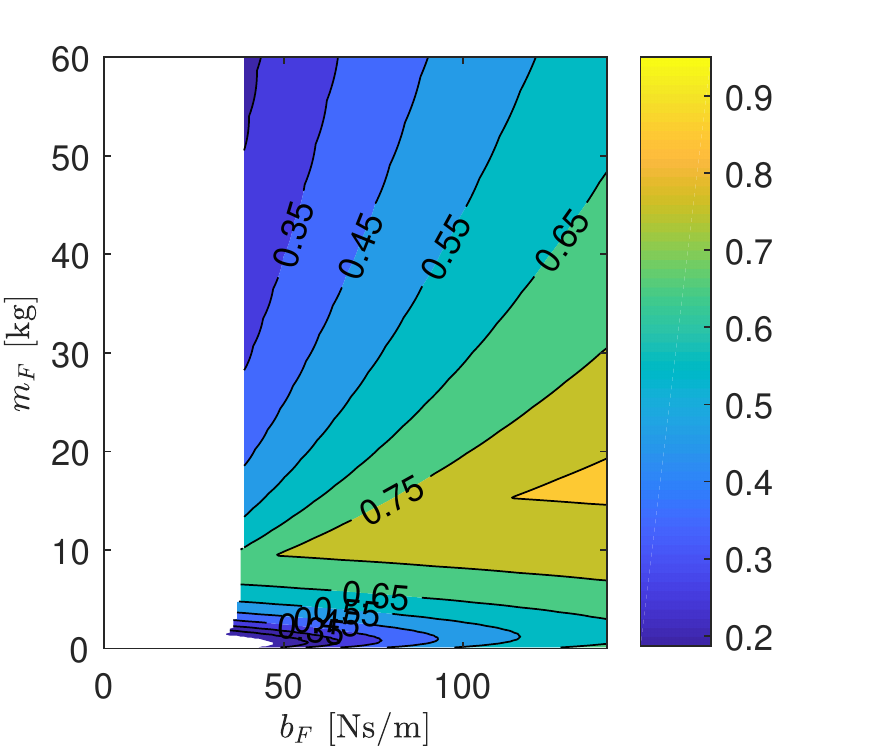}}}
        } \vspace{-0.75\baselineskip}
        \\
        \subfloat[$\alpha=0.7$, $m_{\text{eq}}=0$ kg, \newline $b_{\text{eq}}=0$ Ns/m\label{Fig:rob1_6}]{%
            \resizebox{0.24\columnwidth}{!}{\rotatebox{0}{\includegraphics[trim=0cm 0.05cm 1.05cm 0.2cm, clip=true]{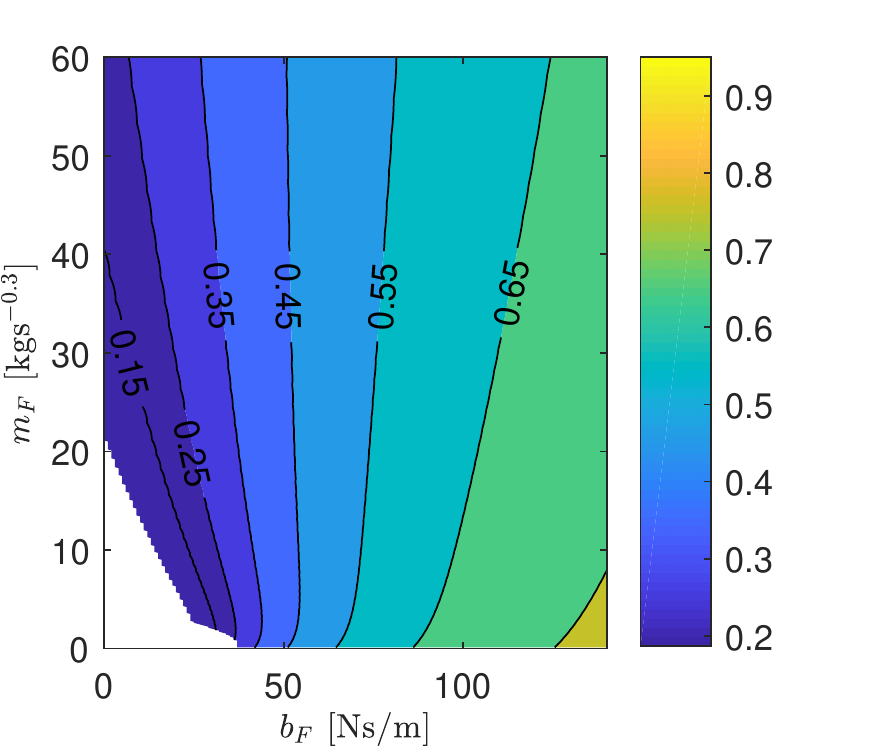}}}
        }
        \hfill
        \subfloat[$\alpha=0.7$, $m_{\text{eq}}=5$ kg, \newline $b_{\text{eq}}=0$ Ns/m\label{Fig:rob7_6}]{%
            \resizebox{0.24\columnwidth}{!}{\rotatebox{0}{\includegraphics[trim=0cm 0.05cm 1.05cm 0.2cm, clip=true]{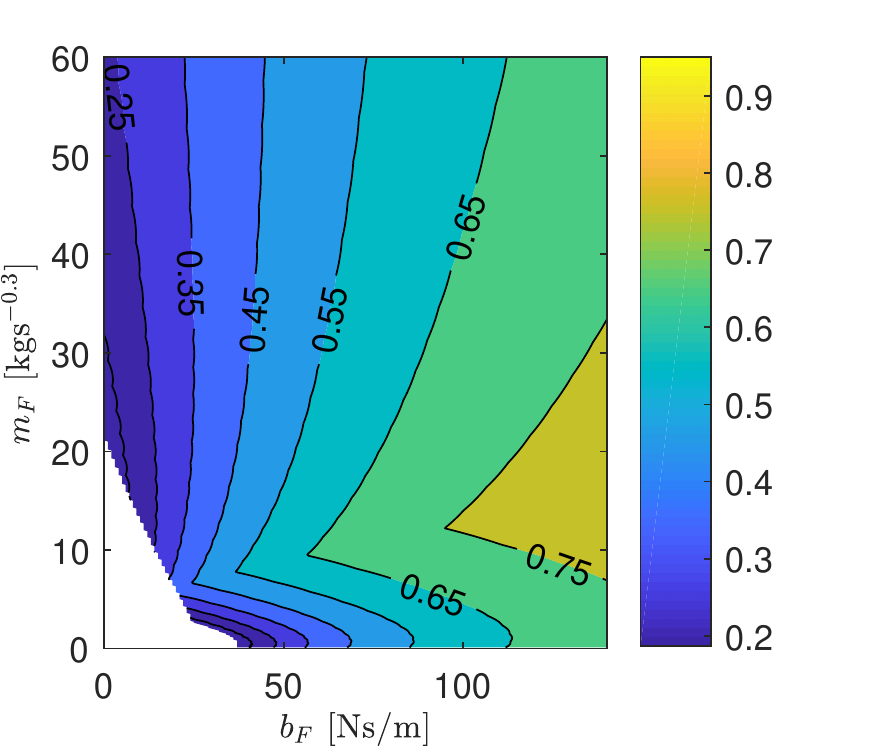}}}
        }
        \hfill
        \subfloat[$\alpha=0.7$, $m_{\text{eq}}=0$ kg, \newline $b_{\text{eq}}=41$ Ns/m\label{Fig:rob4_6}]{%
            \resizebox{0.24\columnwidth}{!}{\rotatebox{0}{\includegraphics[trim=0cm 0.05cm 1.05cm 0.2cm, clip=true]{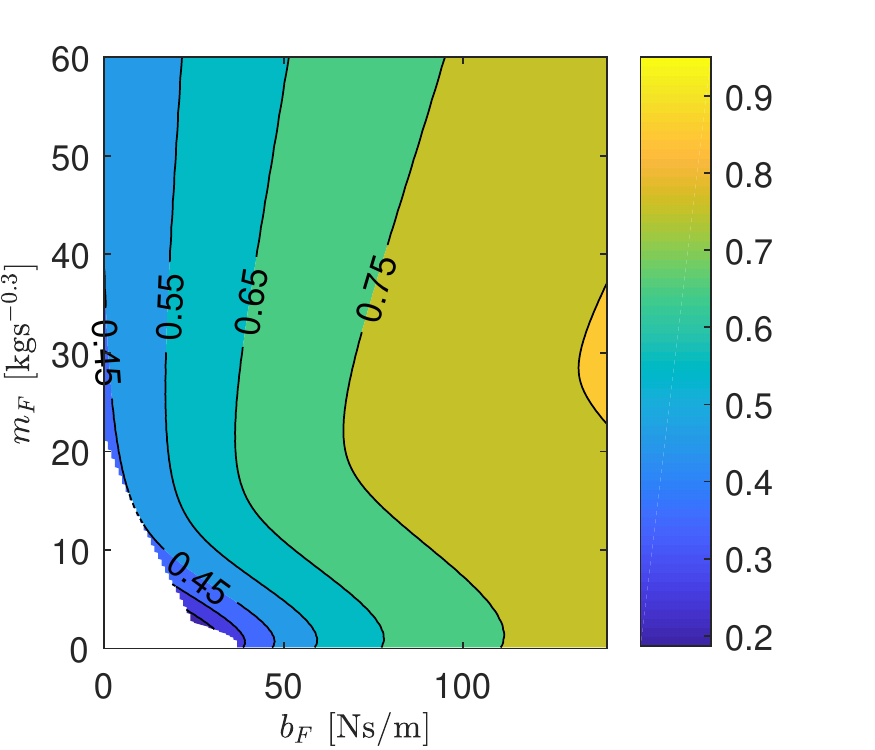}}}
        }
        \hfill
        \subfloat[$\alpha=0.7$, $m_{\text{eq}}=5$ kg, \newline $b_{\text{eq}}=41$ Ns/m\label{Fig:rob4_6}]{%
            \resizebox{0.24\columnwidth}{!}{\rotatebox{0}{\includegraphics[trim=0cm 0.05cm 1.05cm 0.2cm, clip=true]{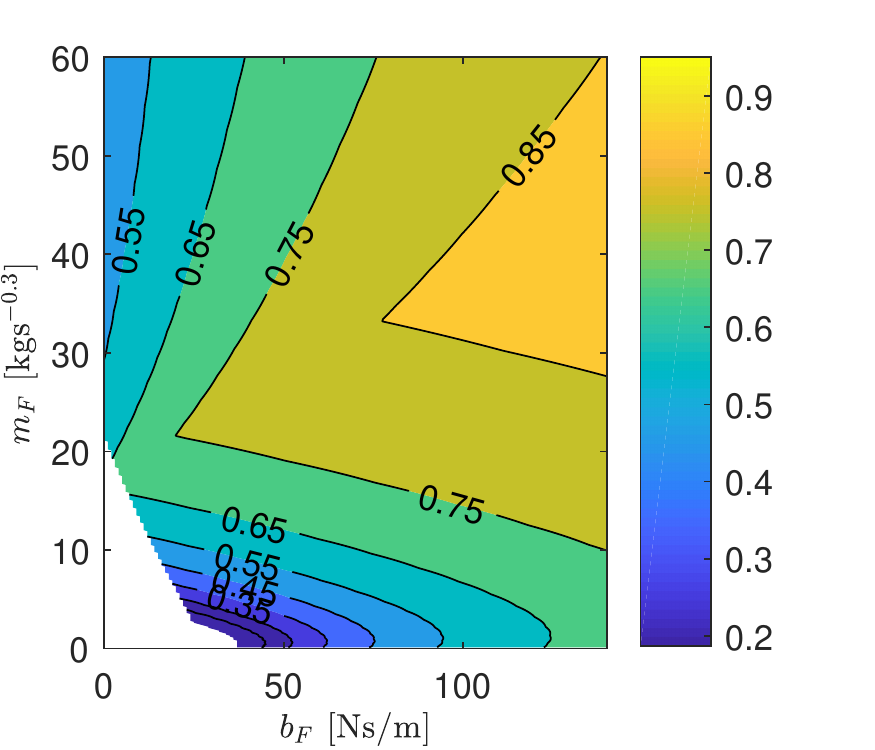}}}
        }	 \vspace{-0.75\baselineskip}
        \\
        \subfloat[$\alpha=0.4$, $m_{\text{eq}}=0$ kg, \newline $b_{\text{eq}}=0$ Ns/m\label{Fig:rob1_6}]{%
            \resizebox{0.24\columnwidth}{!}{\rotatebox{0}{\includegraphics[trim=0cm 0.05cm 1.05cm 0.2cm, clip=true]{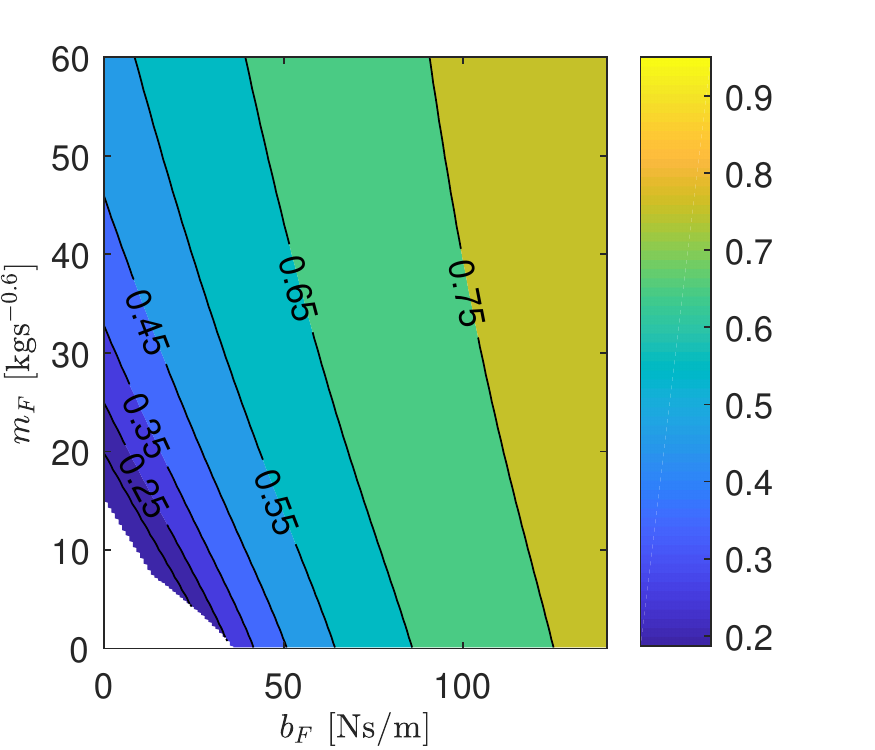}}}
        }
        \hfill
        \subfloat[$\alpha=0.4$, $m_{\text{eq}}=5$ kg, \newline $b_{\text{eq}}=0$ Ns/m\label{Fig:rob7_6}]{%
            \resizebox{0.24\columnwidth}{!}{\rotatebox{0}{\includegraphics[trim=0cm 0.05cm 1.05cm 0.2cm, clip=true]{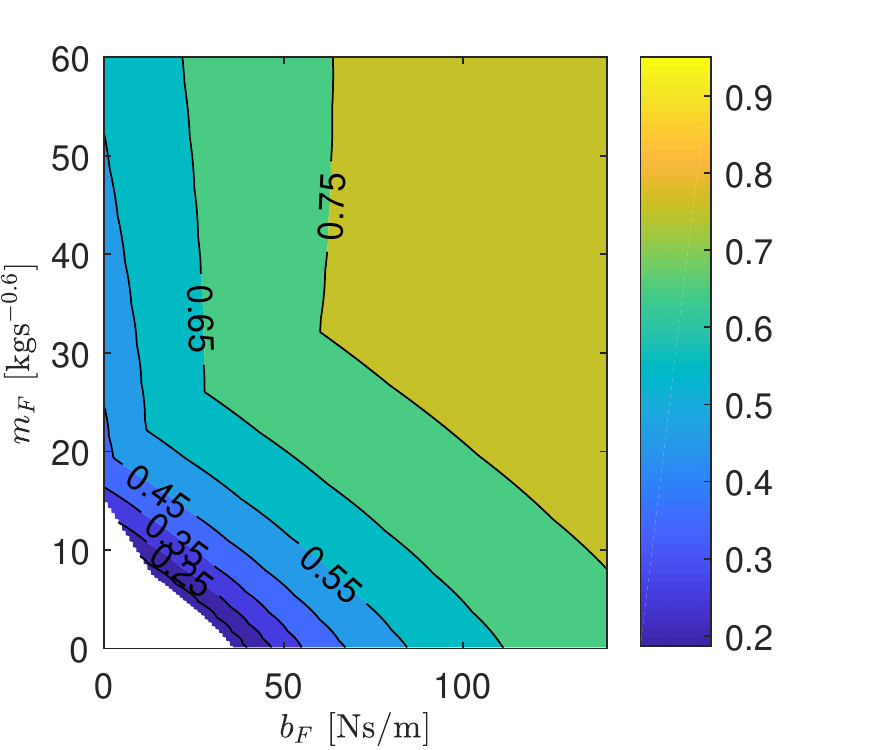}}}
        }
        \hfill
        \subfloat[$\alpha=0.4$, $m_{\text{eq}}=0$ kg, \newline $b_{\text{eq}}=41$ Ns/m\label{Fig:rob4_6}]{%
            \resizebox{0.24\columnwidth}{!}{\rotatebox{0}{\includegraphics[trim=0cm 0.05cm 1.05cm 0.2cm, clip=true]{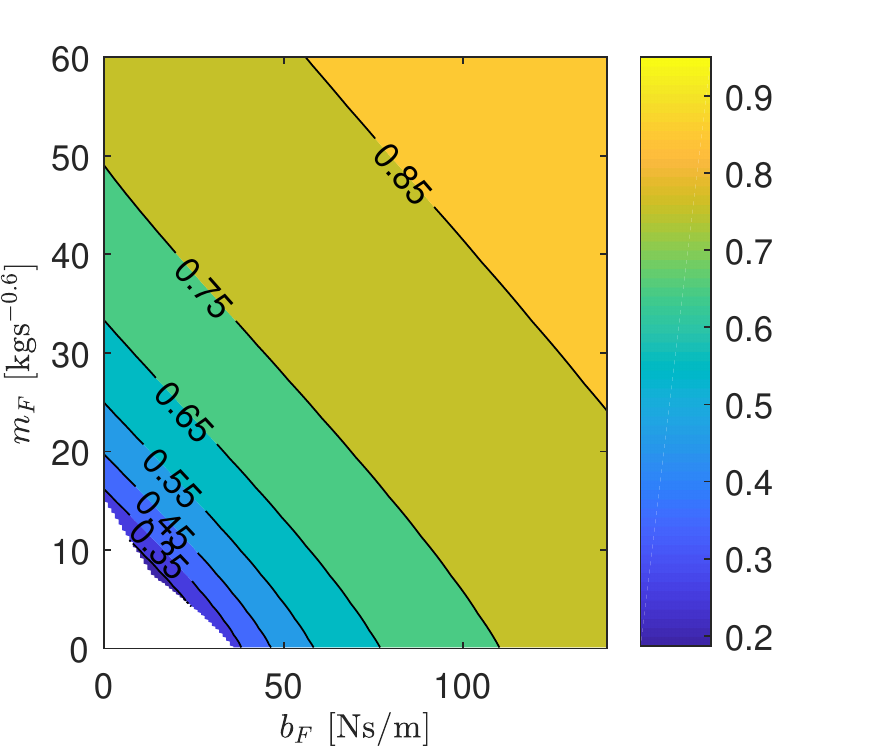}}}
        }
        \hfill
        \subfloat[$\alpha=0.4$, $m_{\text{eq}}=5$ kg, \newline $b_{\text{eq}}=41$ Ns/m\label{Fig:rob4_6}]{%
            \resizebox{0.24\columnwidth}{!}{\rotatebox{0}{\includegraphics[trim=0cm 0.05cm 1.05cm 0.2cm, clip=true]{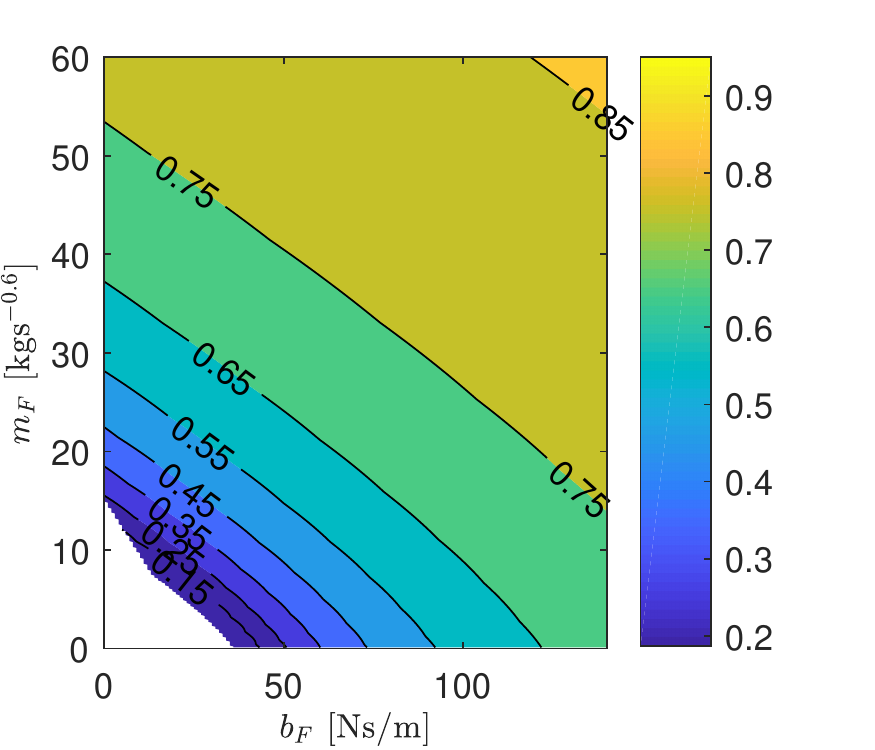}}}
        }
    \end{minipage}
    \vspace{-.75\baselineskip}
    \caption{Stability robustness maps of of the controller in Figure~\ref{Fig:ControlArchitecture} for $k_{\text{eq}}=600$ N/m under IOAC $\alpha=1$, FOAC $\alpha=0.7$, and FOAC $\alpha=0.4$.}
    \vspace{-.5\baselineskip}
    \label{Fig:robMaps}
\end{figure*}
\begin{figure*}[htb]
    \centering
    \begin{minipage}{\textwidth}
        \centering
        \subfloat[$\alpha=1$\label{Fig:rob1}]{%
            \resizebox{0.3\columnwidth}{!}{\rotatebox{0}{\includegraphics[trim=0cm 0.05cm 1.05cm 0.2cm, clip=true]{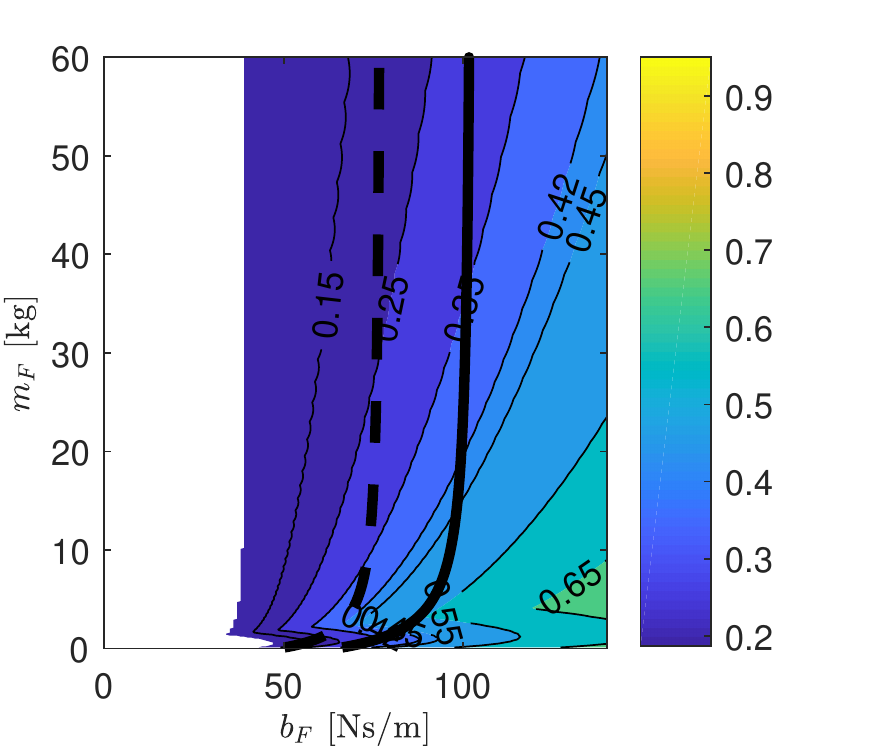}}}
        }
        \hfill
        \subfloat[$\alpha=0.7$\label{Fig:rob7}]{%
            \resizebox{0.3\columnwidth}{!}{\rotatebox{0}{\includegraphics[trim=0cm 0.05cm 1.05cm 0.2cm, clip=true]{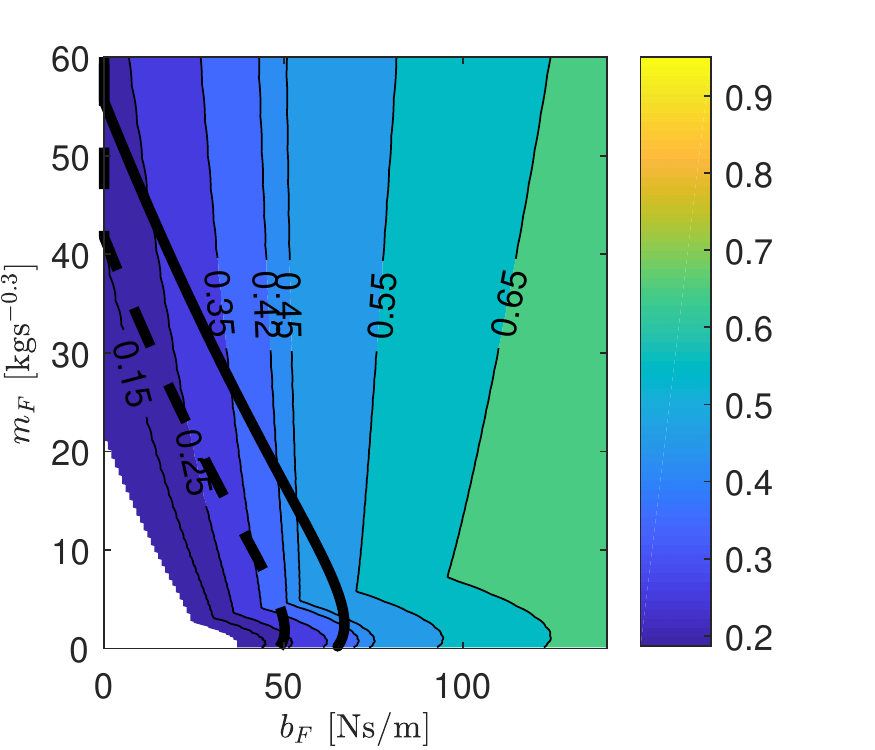}}}
        }
        \hfill
        \subfloat[$\alpha=0.4$\label{Fig:rob4}]{%
            \resizebox{0.3\columnwidth}{!}{\rotatebox{0}{\includegraphics[trim=0cm 0.05cm 1.05cm 0.2cm, clip=true]{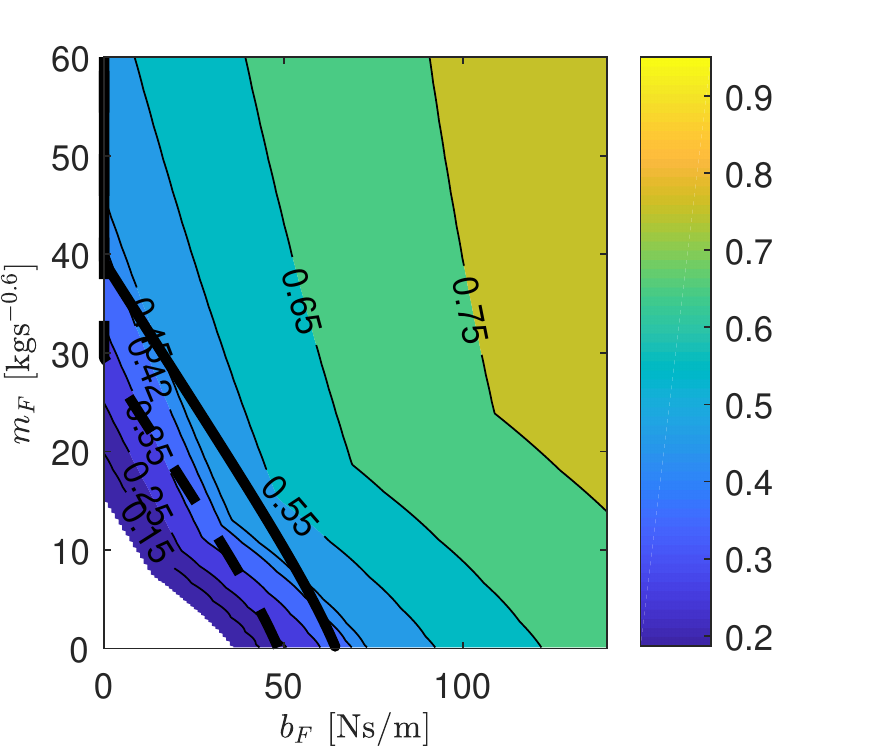}}}
        }
    \end{minipage}
    \vspace{-\baselineskip}
    \caption{Conservative stability robustness maps of the controller   for $k_{\text{eq}}=600$ N/m under (a) IOAC $\alpha=1$, (b) FOAC $\alpha=0.7$, and (c) FOAC $\alpha=0.4$. Solid and dashed thickened black curves show the stability boundaries for $k_{\text{eq}}=1210$ N/m and $k_{\text{eq}}=1610$ N/m, respectively.}
    \label{Fig:robRefinedMaps}  \vspace{-.5\baselineskip}
\end{figure*}

We utilize the vector margin, defined as the inverse of the maximum magnitude of
loop sensitivity function $S(s)$, as the metric of
the stability robustness. For the system in Figure~\ref{Fig:ControlArchitecture}, the sensitivity transfer function is defined as $S(j\omega) = 1/(1+L(j\omega))$, where $L(j\omega)=G(j\omega)Y(j\omega)H(j\omega)Z_{\text{eq}}(j\omega)$ denotes the
loop transfer function.  Then, the vector margin used as the cost function to quantify stability robustness, is defined as
\begin{align}
    \rho =  \frac{1}{\max{(|S(j\omega)|)}}
    \label{Eqn:rho}
\end{align}
\noindent where $\max(|S(j\omega)|)$ represents the maximum magnitude of the
loop sensitivity function. From this equation, one can note that the larger the
loop gain, the more robust the system.

Three sets of robustness maps
for $\alpha \in \{1, 0.7, 0.4\}$ are generated as presented at different rows of Figure~\ref{Fig:robMaps}. Note that, these robustness maps are only constructed for the stable sets of controller parameters for the first pHRI
scenario (S1), as this case is treated as a baseline for other scenarios (S2 and S3) that involve interactions with an environment.

To generate \emph{worst-case} stability robustness maps for each integration order $\alpha$, four different robustness margins are calculated for each stable parameter set: one for each combination of extreme
values of equivalent mass $m_{\text{eq}}$ and damping $b_{\text{eq}}$. For any
given controller parameter set, the minimum of the robustness margins calculated for
these four combinations of extreme values of $m_{\text{eq}}$ and $b_{\text{eq}}$
is taken as the \emph{conservative} robustness margin, so that the worst-case uncertainties in $m_{\text{eq}}$ and $b_{\text{eq}}$ are accounted for in the conservative robustness maps presented in Figure~\ref{Fig:robRefinedMaps}.

In Figure~\ref{Fig:robRefinedMaps}, the relation between the stability robustness margin and equivalent stiffness is depicted. These plots can be utilized to determine how much increase in equivalent stiffness can be tolerated in scenarios S2 and S3. In particular, it can be deduced that a maximum equivalent stiffness increase of 610~N/m and 1010~N/m can be tolerated for S2 and S3, respectively; that is, the maximum expected equivalent stiffness can be set to $k_{\text{eq}} = 1210$~N/m and $k_{\text{eq}} = 1610$~N/m for S2 and S3, respectively, without sacrificing stability.

The stability boundaries for $k_{\text{eq}} = 1210$~N/m and
$k_{\text{eq}} = 1610$~N/m are computed by
employing the principles in~\cite{Aydin2018} and superimposed on top of the stability
robustness maps in Figure~\ref{Fig:robRefinedMaps}. One can observe from
Figure~\ref{Fig:robRefinedMaps}a that among the controller
parameter sets laying on the stability boundary for $k_{\text{eq}} = 1210$~N/m
(resp. $k_{\text{eq}} = 1610$ N/m) under IOAC, the most robust one possesses
a stability robustness of $\rho = 0.42$ (resp. $\rho = 0.55$). In other words,
any controller parameter set of IOAC that results in $\rho = 0.42$ (resp. $\rho = 0.55$)
ensures the coupled stability for S2 (resp. S3),
under the impedance bounds defined for the human and the environment.

Any controller parameter set of FOAC ($\alpha = $0.7 and 0.4 as presented in Figures~\ref{Fig:robRefinedMaps}b and~\ref{Fig:robRefinedMaps}c) that results in the same robustness margin of $\rho = 0.42$ ($\rho = 0.55$) also ensures coupled stability for S2 (resp. S3), as equivalent stiffness of $k_{\text{eq}} = 1210$~N/m (resp. $k_{\text{eq}} = 1610$ N/m) can be tolerated. Note that the controller parameters for FOAC
having the stability robustness margin of $\rho = 0.42$ ($\rho = 0.55$) can tolerate even larger
equivalent stiffness values as they lay further away from
the stability boundary.

\medskip
\textbf{Step 4} -- \emph{Pareto Optimization}
\label{SubSec:CaseStudyParetoOptimization}

We aim to minimize the transparency cost function $C$, while maximizing the
value of the stability robustness margin $\rho$. To eliminate the scale difference between
the two metrics, we first normalize each one with their maximum value, where the maximum values are acquired from the transparency maps (Figure~\ref{Fig:transMaps}) and the stability robustness maps
(Figure~\ref{Fig:robRefinedMaps}). Then, we form the final objective function as their convex combination according to
\begin{align}
J = w\ C_n+(1-w)(-\rho_n)
\label{Eqn:Jobjective}
\end{align}
\noindent where  $w\in[0,1]$ represents
the weights, while $C_n$ and $\rho_n$ denote the normalized values of  transparency cost function and stability robustness margin, respectively. Each value of the weight parameter $w$ results a different
optimization problem, corresponding to various prioritization of the cost functions. If the problem is solved for  all possible weights $w\in[0,1]$,  then resulting set of optimal solutions yield the Pareto front for the multi-criteria optimization problem.

\begin{figure}[b!]
    \centering \vspace{-1.5\baselineskip}
    \resizebox{.9\columnwidth}{!}{\rotatebox{0}{\includegraphics[clip=true]{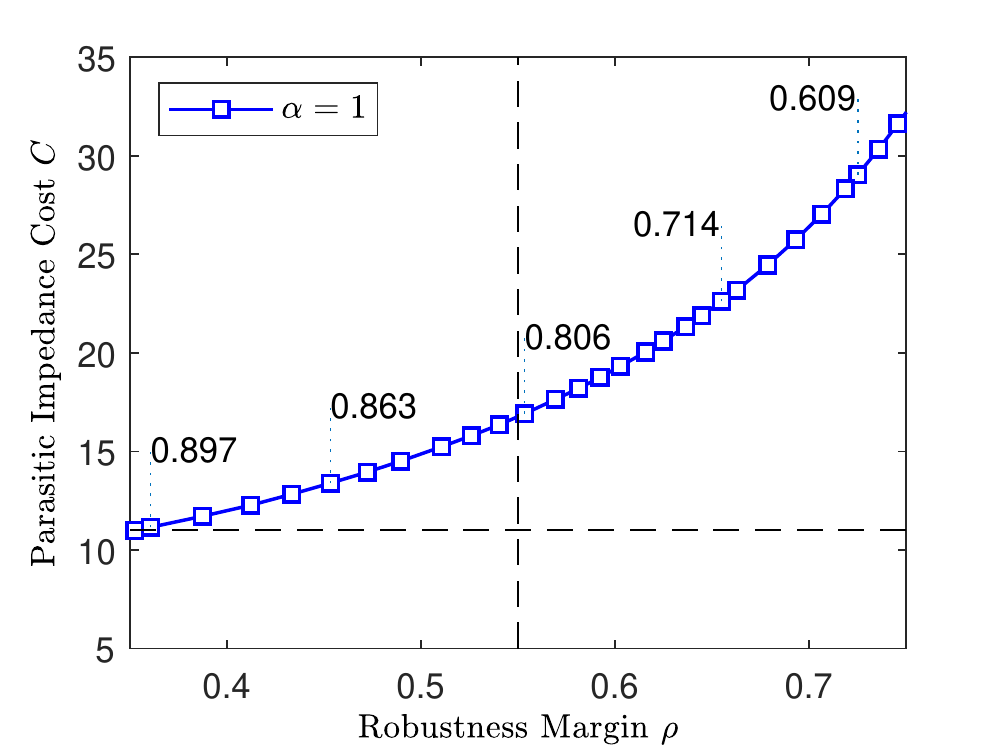}}}
    \vspace{-.8\baselineskip}
    \caption[]{Pareto front curve consisting of non-dominated solutions denoted by empty squares that are computed using multi-criteria optimization for IOAC ($\alpha=$1). The vertical dotted-lines are for connecting the weights corresponding to sample solutions. The horizontal dashed-line denotes a sample
        upper bound for parasitic impedance at $C=11$. The vertical
        dashed-line represents the stability robustness of $\rho=0.55$.
        \label{Fig:paretoIOAC}}
\end{figure}

During our implementation, the weight parameter $w$ is scanned with equally spaced increments of 0.001 to construct the Pareto front curve. Figure~\ref{Fig:paretoIOAC} presents the Pareto front curve for the integration order $\alpha=1$. A close inspection of Figure~\ref{Fig:paretoIOAC} reveals the trade-off between
the selected cost functions. The maximum robustness margin is achieved
when the parasitic impedance is at its maximum value, resulting in a poor
transparency performance. On the contrary, the parasitic impedance attains its minimum, yielding to the highest transparency, when the robustness margin is at its minimum, leading to the lowest stability robustness.

Prioritization of objectives without having a thorough understanding on the trade-off characteristics may lead to conservative designs, whereas Pareto front curve characterizes the trade-off between objectives thoroughly so that decision making can be carried out after inspection of all possible optimal solutions. In Figure~\ref{Fig:paretoIOAC}, the weights for several solutions on the Pareto front curve are displayed. Note that choosing a lower weight prioritizes the stability robustness, while a higher one promotes transparency. To obtain an optimal solution by the single criterion optimization (under scalarization) requires a proper choice for the weight, which is unknown a priori. Inspecting Pareto front curve reveals that such a weight cannot be determined by intuition alone, and a thorough analysis is required so that the designer can make an informed decision.

\subsection{Design Selection}
\label{SubSec:CaseStudyDesignSelection}

An optimal solution is selected from the Pareto front curve in three steps.
Considering the pHRI scenarios discussed in Section~\ref{SubSec:pHRITasks}, we
illustrate by examples how to i) impose possible constraints on the
optimization objectives, ii) consider secondary design criteria, and iii) select an optimal design by
deciding on an optimal solution on the Pareto front curve.

\subsubsection{Constraints on the Optimization Objectives}
\label{SubSubSec:CaseStudyConstraints}

Certain constraints may be imposed on the optimization cost functions to ensure some level of performance of a given pHRI task. These constraints need not to be assigned prior to optimization and can be determined after inspecting the Pareto front curve.

Let us consider the Pareto front curve presented in Figure~\ref{Fig:paretoIOAC}. After inspecting the plot, the designer may choose to impose an upper bound of $C<11$ on the parasitic impedance, as marked by the horizontal dashed-line. This line intersects the Pareto front
curve at a robustness margin of $\rho=0.35$. On the other hand, for another
pHRI task, stability robustness may be more critical than the parasitic impedance, if
the environment and/or human arm impedance are prone to significant change. For
example, inspecting Figure~\ref{Fig:robRefinedMaps} reveals that any controller
with a stability robustness margin of $\rho=0.55$ can maintain coupled stability up to $k_{\text{eq}}$ = 1610~N/m, the highest allowable stiffness in scenario S3. Prioritizing this stability margin, the designer can introduce the stability robustness margin of $\rho=0.55$ as a lower bound as depicted by the vertical dashed-line in
Figure~\ref{Fig:paretoIOAC}. This choice eliminates all the optimal
solutions with $\rho<0.55$  from any further consideration.

\begin{figure}[t]
    \centering
    \resizebox{0.90\columnwidth}{!}{\rotatebox{0}{\includegraphics[clip=true]{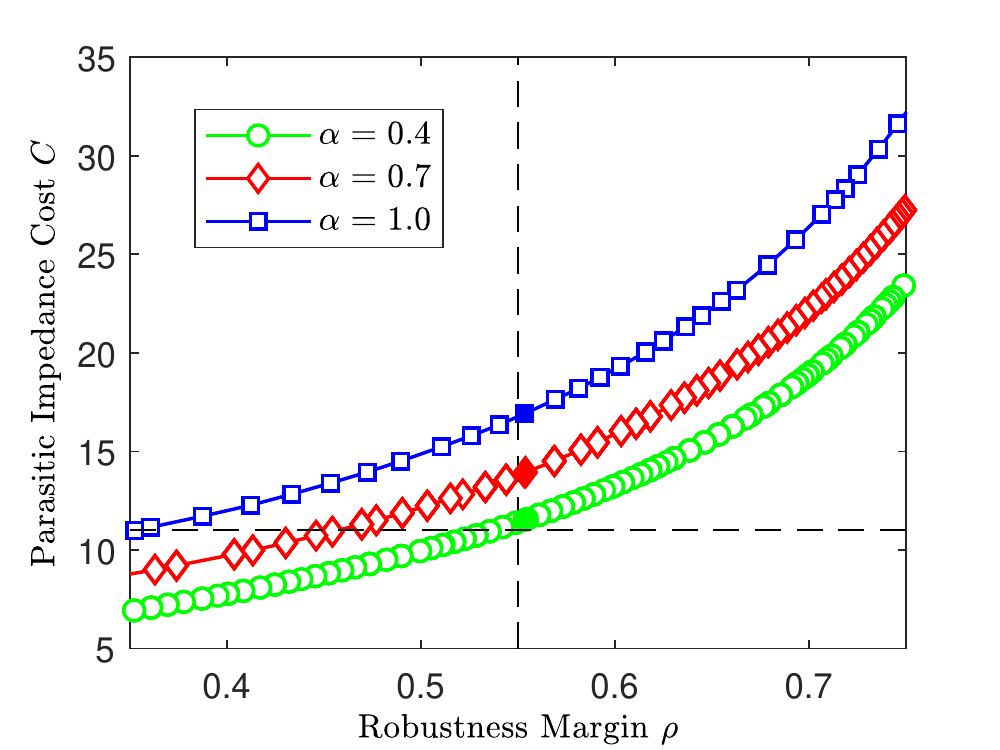}}}
     \vspace{-.8\baselineskip}
    \caption[]{Pareto front curves for IOAC ($\alpha=$1), FOAC ($\alpha=$0.4 and
    0.7). The horizontal dashed-line denotes a sample upper bound of $C\leq11$ for
parasitic impedance. The vertical dashed-line represents a sample lower bound of $\rho\geq0.55$
for the stability robustness. The solutions denoted by filled markers
are the optimal solutions selected, as detailed in
Table~\ref{controlTable55}. \label{Fig:paretocurves}}  \vspace{-.75\baselineskip}
\end{figure}

Same constraints can also be applied to the optimal solutions on Pareto front curves
of other integration orders, presented in Figure~\ref{Fig:paretocurves}. Note
that, when both $C\leq11$ and $\rho\geq 0.55$ constraints are simultaneously considered, there exits
no optimal solution for IOAC, while there are such optimal solutions for FOAC with $\alpha=0.4$ and $\alpha=0.7$.

\subsubsection{Considering Other Design Criteria}

For the case study, stability robustness and  transparency are chosen as the optimization objectives. However, once a thorough understanding of the trade-off between these  objectives is
achieved, the designer is free to introduce new design criteria to further narrow
down the set of optimal solutions.  As an example, noting that intended movements of human arm is
band-limited~\cite{Brooks1990}, we introduce an additional criterion for the pHRI task.
In particular, we introduce a new criteria not captured by the optimization objectives and require the cut-off
frequency of the closed-loop system to be larger than the bandwidth of the human
arm's intended movements.

\begin{figure}[t]
    \centering
    \resizebox{0.90\columnwidth}{!}{\rotatebox{0}{\includegraphics[clip=true]{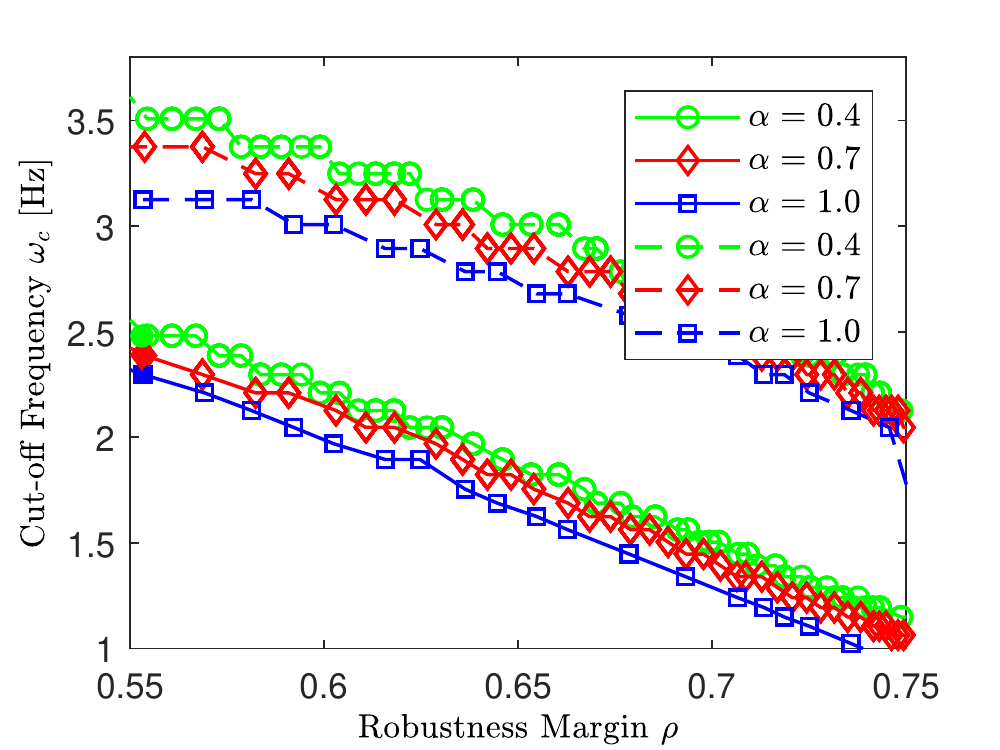}}}
     \vspace{-.8\baselineskip}
    \caption[]{The cut-off frequency $\omega_c$ of $T_\text{disp}(s)$ as a
    function of $\rho$ for the optimal solutions with $\rho \geq 0.55$.
Solid and dashed lines are for $k_e = 610$ and $k_e=1010$ N/m, respectively. The
solutions represented by filled markers denote the  optimal solutions selected,
as detailed in Table~\ref{controlTable55}.\label{Fig:paretofreqs}}  \vspace{-.25\baselineskip}
\end{figure}

During the execution of a pHRI task, the user physically interacts
with a robot and environment whose reflected impedance is characterized by
$Z_\text{disp}(s)=F_h(s)/V(s)$, as in~\eqref{Eqn:closedLoopImpedance}. To study the cut-off frequency
of this closed-loop system, the transfer function $T_\text{disp}(s) =
X(s)/F_h(s)$ is calculated using the reflected impedance and this cut-off frequency
$\omega_c$ is computed for all the remaining
solutions on the Pareto front. In Figure~\ref{Fig:paretofreqs}, these cut-off frequencies $\omega_c$ are presented as a function of stability robustness $\rho$.

\subsubsection{Deciding on an Optimal Controller}
\label{SubSubSec:CaseStudyDecidingOptimalController}

As $k_e=610$ N/m is the lower bound on the environment stiffness for scenario S3, the cut-off frequencies $\omega_c$ computed for this stiffness level are considered for the design, noting that, an
increase in the environment stiffness $k_e$ also increases $\omega_c$
(Figure~\ref{Fig:paretofreqs}). Given that the bandwidth of the intended movements of
human arm is around 2~Hz~\cite{Dimeas2016,Brooks1990}, with a tolerance of 15$\%$, the solutions leading to
$\omega_c < 2.3$ Hz are also eliminated from any further consideration.

An optimal solution that
satisfies $\omega_c \geq 2.3$ was selected as the final design for each
integration order (see Table~\ref{Tab:controlTableCutoff}). We observe that FOAC
(with $\alpha=0.4$ and $\alpha=0.7$) is a better choice than IOAC (with $\alpha=1$) since FOAC
results in higher stability robustness $\rho$ and lower parasitic impedance $C$, leading to more robust and transparent design as presented in Table~\ref{Tab:controlTableCutoff}.

An important advantage of Pareto methods is that designer only needs to decide on an optimal solution and the trade-off between objectives after obtaining the Pareto front. The weighting parameter, $w$, for each sample design is given in Table~\ref{Tab:controlTableCutoff}. Note that values of $w$ for each controller is different and these values cannot be easily determined by intuition a priori. A thorough knowledge on the trade-off characteristics is needed before deciding on these values.

\begin{table}[]
    \centering
    \caption{The optimal controller parameters, cost function values, and optimization
weights for $\omega_c \geq 2.3$ Hz and $\rho \geq 0.55$. \label{Tab:controlTableCutoff}}
 \vspace{-.8\baselineskip}
    \begin{tabular}{cccccc}
        \toprule
        $\alpha$   & $m_F$ [kgs$^{\alpha-1}]$    & $b_F$ [Ns/m]  & $\rho$  & $C$ & $w$    \\
        \midrule
        1   & 3.2  & 90 & 0.553 & 16.9 & 0.796\\
        0.7 & 6.0  & 74 & 0.568 & 14.5 & 0.755\\
        0.4 & 16.7 & 56 & 0.594 & 13.0 & 0.737\\
        \bottomrule
    \end{tabular}  \vspace{-.8\baselineskip}
\end{table}

\subsection{Comparison of Interaction Controllers}
\label{SubSec:CaseStudyControllerComparison}

Using Pareto optimization approach not only allows the designer to make an
informed decision by inspecting all optimum solutions and studying the trade-off between the objectives, but
also enables fair comparisons among various interaction controllers, possibly having
different structures. 

A close inspection of the Pareto front curves in Figure~\ref{Fig:paretocurves} reveals an interesting and important result: The Pareto front curve for the IOAC is completely dominated by those of the FOACs. In other words, parasitic impedance under IOAC is always larger than those of FOACs for the same robustness margin;
hence, FOAC allows a better compromise between the stability robustness and
transparency. Furthermore, it can be easily observed that the Pareto front curve for
integration order $\alpha=0.4$ is superior to the others. Hence, the best compromise between stability robustness and transparency is obtained under $\alpha=0.4$ among the controllers considered in this study.

\begin{table}[t!]
    \centering 
    \caption{Parameters and the parasitic impedance $C$ of the optimal controllers for scenario S3.}  \vspace{-.8\baselineskip}
    \begin{tabular}{ccccccc}
        \toprule
        $\alpha$   & $m_F$ [kgs$^{\alpha-1}]$    & $b_F$ [Ns/m]  & $C$ & $w$ \\
        \midrule
        1   & 3.2  & 90 & 16.9 & 0.796 \\
        0.7 & 5.8  & 71 & 13.9 & 0.770 \\
        0.4 & 15.4 & 49 & 11.5 & 0.776 \\
        \bottomrule
    \end{tabular}  \vspace{-.8\baselineskip}
    \label{controlTable55}
\end{table}

\section{A Controlled pHRI Experiment: Continuous Contact with Environment}
\label{Sec:Experiments}

The controllers tabulated in Table~\ref{controlTable55} were implemented for a pHRI experiment and a thorough comparison is reported in this section. In particular, optimal FOAC with integration orders of $\alpha=0.4$ and $\alpha=0.7$ that have the same stability robustness as the optimal IOAC ($\rho=0.553$)   were designed and implemented, to experimentally compare their transparency performance.  The task involved contact interactions with a nonlinear environment formed by two linear springs that engage at different positions of the end-effector.

\subsection{Experimental Procedure}
\label{SubSec:SpringExpProcedure}
\begin{figure}[b!]
    \centering
    \resizebox{0.8\columnwidth}{!}{\rotatebox{0}{\includegraphics[clip=true]{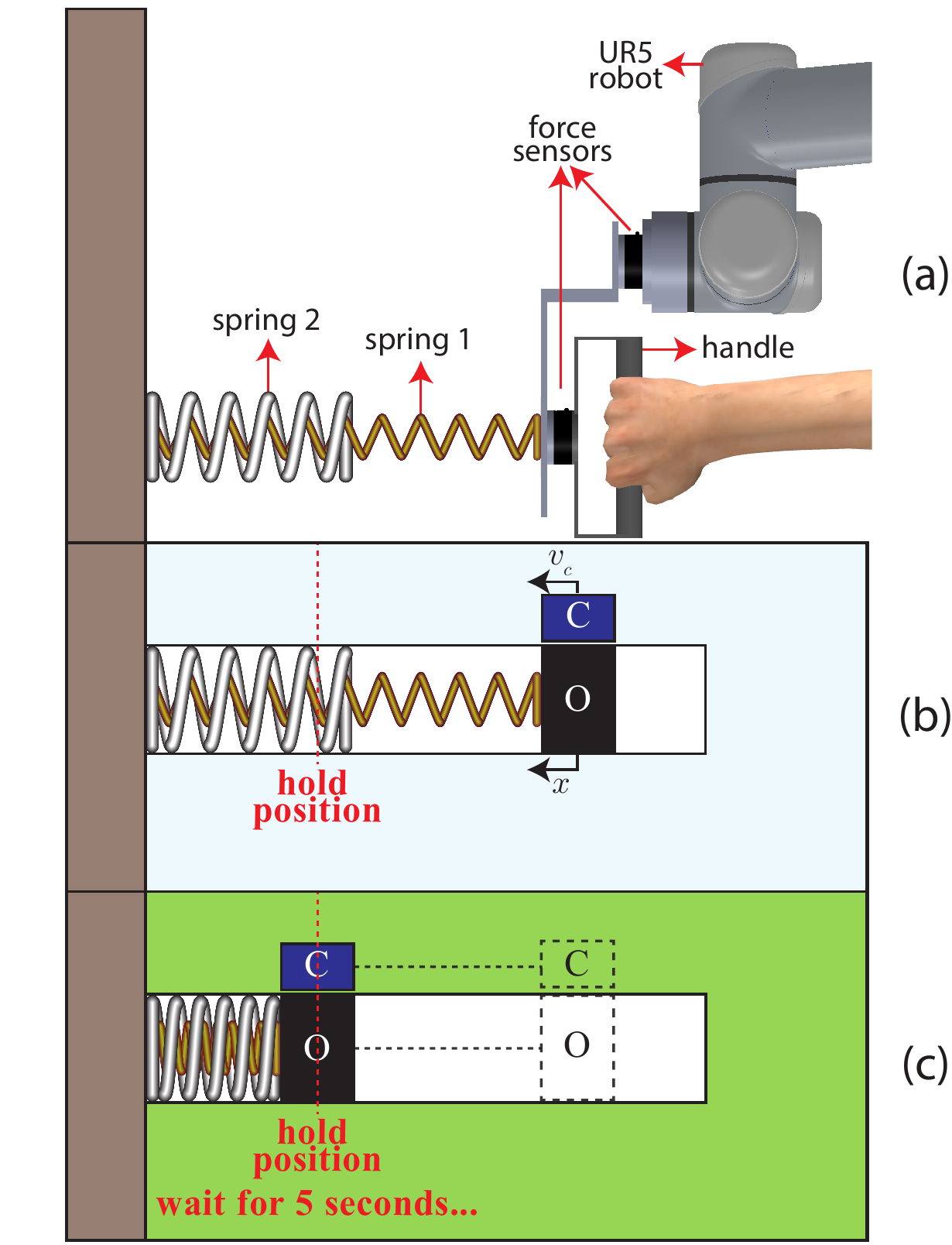}}}
    \vspace{-.7\baselineskip}
    \caption[]{The experimental setup (a) used, and the visual feedback (b,c) displayed to the subjects during the experiments.\label{fig:expSetup}} 
\end{figure}

In the experimental setup, human operator manipulates the robot to interact with
an environment, where the environment stiffness is varied based on the depth of compression. In particular,  two different springs are used as depicted in Figure~\ref{fig:expSetup}. Spring~1 with $k_1 = 610\text{ N/m}$ is connected between a fixed support and the end-effector of
a UR5 robot (Universal Robots Inc.), while Spring 2 with $k_2 = 400\text{
N/m}$ becomes active when the depth of compression is larger than
\SI{45}{\milli\meter}. Hence, the environment stiffness ($k_e$) may change
abruptly from \SIrange{610}{1010}{N/m} during the interaction.

The subject grasps the handle to guide the end-effector and compresses the
springs. The springs simply represent a layered environment in which the robot
guided by a human operator is in continuous contact with, as in drilling a wall or
inserting a needle into soft tissue. In the experiments, the subjects are asked
to compress the springs up to a depth of \SI{55}{\milli\meter}, called as the ramp phase (Figure~\ref{fig:expSetup}b), and then, hold it at that position for 5~s,
called as the hold phase (Figure~\ref{fig:expSetup}c). In order to make the compression rate
equal for each subject, a visual cursor (the dark blue rectangle, labelled as
”C” in Figure~\ref{fig:expSetup}b) moving with a constant speed ($v_c=20$~mm/s)
is displayed on the computer screen. The other details of the experimental
setup is the same as the one reported in~\cite{yusufWHC}.

Ten subjects ($5$ males and $5$
females, average age $= 29\pm6$) participated in this experiment. The subjects
gave informed consent about their participation in the experiment. The
experimental study was approved by the Ethical Committee for Human Participants
of Koc University.

We compared subjects' performance for the controllers tabulated in
Table~\ref{controlTable55}. Hence, three different controllers, IOAC ($\alpha =
1$), FOAC ($\alpha=0.7$), and FOAC ($\alpha = 0.4$) were tested in this
experiment. The task was repeated 10 times for each controller, resulting in
30 ($3\times10$) trials in the experiment, which were displayed to the subjects
in random order. The order was same for each subject. All trials were performed one after the other without any breaks. Prior to the experiment, each subject was given a training session of 15 ($3\times5$) trials to get herself/himself familiar with the setup.

\subsection{Data Collection and Performance Metrics}
\label{SubSec:MetricSection}

During the experiment, the robot constrains the motion of the subjects along a
horizontal line while they compress the springs. The force $F_h$ applied by the
subject is the sum of the forces required for compressing the springs ($F_e$)
and generating the motion trajectory of the robot (i.e. interaction force
$F_{\text{int}}$); $F_h=F_e+F_{\text{int}}$, where $F_e = k_e\Delta x$ ($\Delta
x$ is the amount of compression) is not influenced by the controller.
$F_{\text{int}}$ is the interaction force  measured by a force sensor (Mini40,
ATI Inc.), filtered, and fed back to the admittance controller. The force
applied by the subject $F_h$ (measured by a second sensor, Mini40, ATI Inc.) and the
interaction force $F_{\text{int}}$ are linearly dependent on each other, since
$F_e$ depends on the amount of compression only, and is independent of viscous
and inertial effects coming from the controller and the robot. As the portion of the force applied by the subject to overcome the parasitic impedance of the robot is the
interaction force, by inspecting this force alone, we can compare the
transparency performance of the admittance controller for $\alpha =
1$, $\alpha=0.7$, and $\alpha = 0.4$.

We use average interaction force ($F_{\text{int}}^{\text{ave}}=1/(t_e-t_b)
\int_{t_b}^{t_e}\left |F_{\text{int}}(t)\right |dt$) to quantify the interaction
performance under different controllers, and average force applied by human
subjects ($F_{h}^{\text{ave}}=1/(t_e-t_b) \int_{t_b}^{t_e}\left |F_{h}(t)\right
|dt$) to investigate the human effort, where, $t_b$ and $t_e$ are the beginning
and ending times of a phase (ramp or hold), respectively. We also inspect the
energy consumed to overcome the parasitic impedance of the robot
($E_{\text{int}}(t) = \int_{t_i}^{t}F_{\text{int}}(t)v(t)dt$) as a function of
time $t$, and then use it for estimating the total energy consumption
($E^{\text{tot}}_{\text{int}} = E_{\text{int}}(t_f)$), where $t_i$ and $t_f$ are
the beginning and ending times of a trial. Moreover, the total work done by the
human ($W_{h} = \int_{t_i}^{t_f}F_{h}(t)v(t)dt$) is also investigated. In
addition, peak amplitude of oscillations in interaction force
$A_{F_{\text{int}}}$, force applied by human $A_{F_h}$, and end-effector
position $A_{P}$ are computed using fast Fourier transform (FFT) analysis as measures to compare the robustness under each controller.

\subsection{Data Analysis}
\label{SubSec:DataAnalysis}

\begin{figure}[b!]
    \centering
    \resizebox{0.90\columnwidth}{!}{\rotatebox{0}{\includegraphics[clip=true]{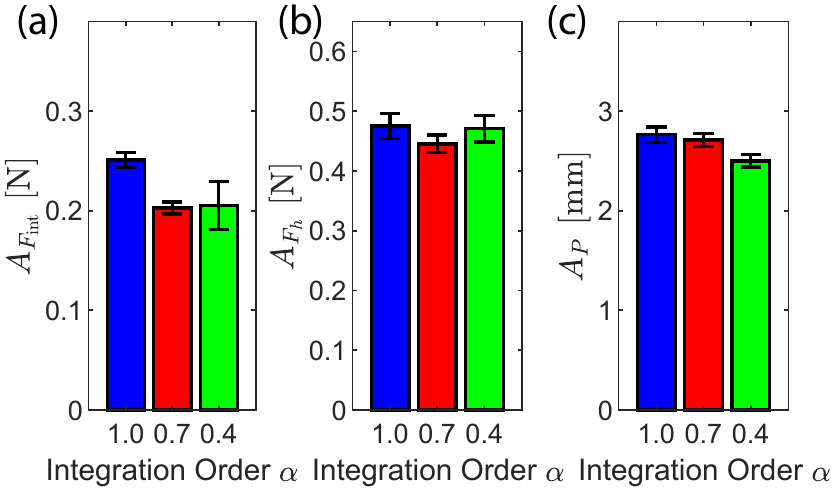}}}
    \vspace{-.8\baselineskip}
    \caption[]{The means and the standard errors of means of normalized
    performance metrics; peak amplitude of oscillations in (a) force applied by
the subjects, $A_{F_h}$, (b) interaction force, $A_{F_{\text{int}}}$, and (c)
end-effector position, $A_P$, during the hold phase. \label{fig:amps}} 
\end{figure}

For each subject, the performance metrics $E^{\text{tot}}_{\text{int}}$ and
$W_{h}$ were calculated for the entire duration,  $F_{\text{int}}^{\text{ave}}$
and $F_{h}^{\text{ave}}$ were computed for the ramp phase, $A_{F_{\text{int}}}$,
$A_{F_h}$, and $A_P$ were evaluated for the hold phase of all trials and then
the mean values were normalized for the analysis
(Figures~\ref{fig:amps},~\ref{fig:emax}, and~\ref{fig:aveForce}).

We performed
one-way ANOVA to investigate the statistical significance of these results. In
all statistical analyses, a significance level of $p=0.005$ was used for the
null hypothesis.

\subsection{Results}
\label{SubSec:ResultsSpring}

Figures~\ref{fig:amps}a,~\ref{fig:amps}b, and~\ref{fig:amps}c present the average
peak amplitude of oscillations in force applied by the subjects $A_{F_h}$,
interaction force $A_{F_{\text{int}}}$, and end-effector position $A_P$ during
the hold phase, respectively. Although these values were slightly lower for
FOAC, the differences were not significant.

\begin{figure}[t!]
    \centering\vspace{-.5\baselineskip}
    \resizebox{0.80\columnwidth}{!}{\rotatebox{0}{\includegraphics[clip=true]{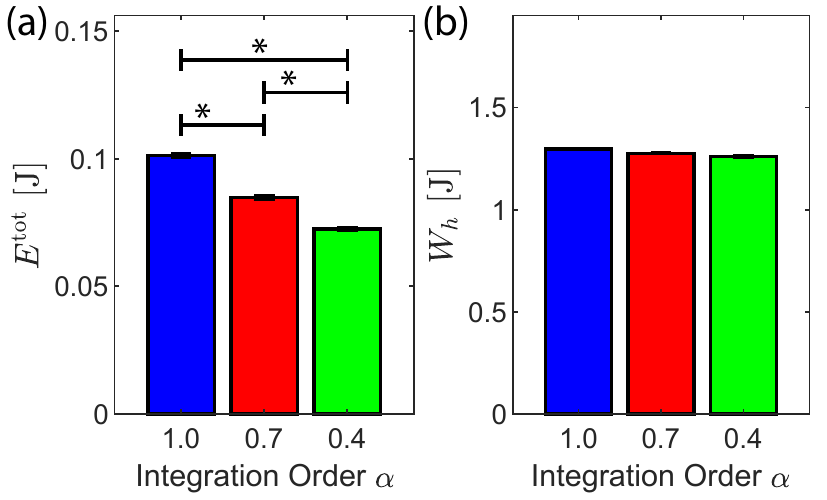}}}
    \vspace{-.8\baselineskip}
    \caption[]{The means and the standard errors of means of normalized
    performance metrics; (a) average energy consumed to overcome the parasitic
impedance of the robot, $E^{\text{tot}}_{\text{int}}$, and (b) the work done by
the subjects, $W_{h}$ (Horizontal bars with * on top indicate statistical
significance between the results of the two corresponding
controllers).\label{fig:emax}}\vspace{-.5\baselineskip}
\end{figure}

\begin{figure}[b!]
    \centering \vspace{-.8\baselineskip}
    \resizebox{0.80\columnwidth}{!}{\rotatebox{0}{\includegraphics[clip=true]{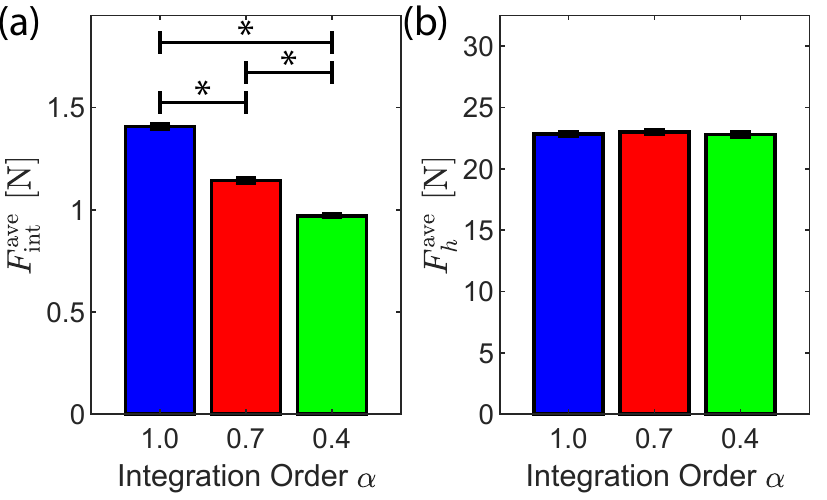}}}
    \vspace{-.8\baselineskip}
    \caption[]{The means and the standard errors of means of normalized
    performance metrics; (a) average interaction force,
$F_{\text{int}}^{\text{ave}}$, and (b) force applied by the subjects,
$F_h^{\text{ave}}$, during the ramp phase (Horizontal bars with * on top
indicate statistical significance between the results of the two corresponding
controllers).\label{fig:aveForce}}
\end{figure}

Figures~\ref{fig:emax}a and~\ref{fig:emax}b illustrate the average energy
consumed to overcome the parasitic impedance of the robot
$E^{\text{tot}}_{\text{int}}$, and the work done by the subjects $W_{h}$,
respectively. We observed a statistically significant effect of controller on
$E^{\text{tot}}_{\text{int}}$. Specifically, it was significantly lower under
FOAC than that of IOAC. Moreover, it was the lowest under FOAC ($\alpha=0.4$).
In addition, FOAC  slightly reduced the work done by the subjects.

Average interaction force $F_{\text{int}}^{\text{ave}}$, and force applied by
the subjects $F_h^{\text{ave}}$ during the ramp phase were depicted in
Figures~\ref{fig:aveForce}a and~\ref{fig:aveForce}b, respectively. We observed a
statistically significant effect of controller on $F_{\text{int}}^{\text{ave}}$.
Specifically, $F_{\text{int}}^{\text{ave}}$ was significantly lower under FOAC
than that of IOAC. Moreover, $F_{\text{int}}^{\text{ave}}$ was the lowest under
FOAC ($\alpha=0.4$).

\subsection{Discussion}
\label{SubSec:ExperimentalResultDiscussion}

\begin{figure}[t!]
    \centering
    \resizebox{0.70\columnwidth}{!}{\rotatebox{0}{\includegraphics[clip=true]{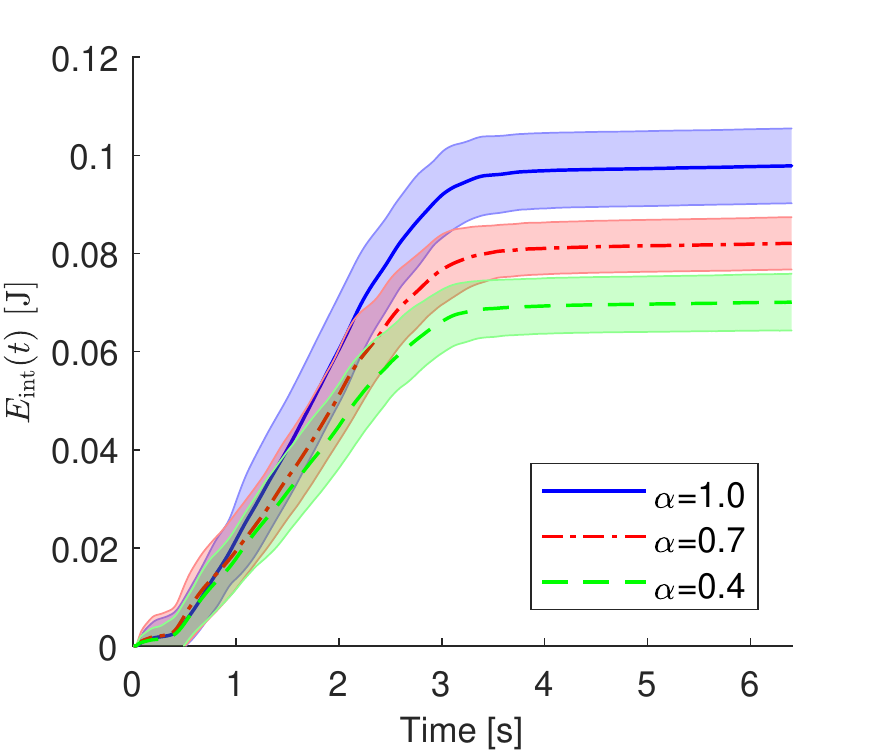}}}
     \vspace{-.8\baselineskip}
    \caption[]{Energy consumed to overcome the parasitic impedance of the robot as a function of time, $E_{\text{int}}(t)$.\label{fig:energy}}\vspace{-\baselineskip}
\end{figure}

We compared the performance of FOAC with that of IOAC for the pHRI task of compressing the two-stage spring system. Optimal admittance controllers with the same level of stability robustness were tested in our experiment and their transparency
performances were compared.

None of the subjects experienced any instability while interacting with the
robot during the experiments. The fact that no instability was observed provides evidence that each controller displayed sufficient level of stability robustness, as they were designed to tolerate even the highest equivalent stiffness during the task.

In order to compare the energy dissipation characteristics of these controllers,
the amplitude of oscillations in force applied by the subjects, $A_{F_h}$,
interaction force $A_{F_{\text{int}}}$, and end-effector position of the robot,
$A_P$, were used. We did not expect a significant difference between these
values during the hold phase (Figure~\ref{fig:amps}), where the equivalent
stiffness is the highest. Since each controller was designed to maintain the
robust stability up to $k_{\text{eq}} = 1610$ N/m, their energy dissipation
capacities should be similar.

We expected that the energy consumed to overcome the parasitic impedance of the
robotic system during the ramp phase would be significantly lower under FOAC for
$\alpha = 0.4$, as it was found to yield the lowest parasitic impedance in
our computational analysis as presented in Table~\ref{controlTable55}. Indeed, the parasitic impedance was lower
under FOAC in our experimental study as depicted in Figures~\ref{fig:emax}a
and~\ref{fig:energy}. Furthermore, average interaction force during the ramp
phase was reduced as the integration order was lowered, as can be observed in Figure~\ref{fig:aveForce}a.
The fact that both the energy consumption and the force to overcome the
parasitic impedance were lower under FOAC revealed that a higher transparency was
achieved with this controller. On the other hand, the difference between IOAC and FOAC in terms of
the average force applied (Figure~\ref{fig:aveForce}b) and the work done by the
subjects (Figure~\ref{fig:emax}b) was not significant during the ramp phase
since compressing the springs required a significant portion of the total energy
consumed by the subjects during the task.

\vspace{-0.5\baselineskip}
\section{Discussion and Conclusion}
\label{Sec:Conclusion}

In this study, we presented a  multi-criteria optimization framework for
synthesizing interaction controllers. In particular, we investigated the
trade-off between transparency and stability robustness in pHRI systems and
proposed a computational approach for optimal design of interaction controller.

Our computational approach resembles to the complementary stability analysis suggested in~\cite{Buerger2007}. However, while their method relies on prioritization of
one objective over another and imposing stability as a constraint, our approach advocates computation of the Pareto front that allows the designer to make an informed decision by studying all optimal solutions.

Pareto front solution not only can be used to select an optimal solution for the task at hand, but the same Pareto front may be utilized to select optimal controllers when the task changes, without a need for re-run of the analysis. For instance, in case the same pHRI system is to be used for the execution of another task where the environmental stiffness is higher than that of the previous task, the designer can select a new design on the Pareto front curve with a higher stability robustness to accommodate this change, as all possible optimal solutions are already available on the Pareto front.

The proposed design framework not only enables to optimize the controller parameters for
the best trade-off performance, but also allows for interaction controllers
with different structures to be compared rigorously. Note that, for interaction controllers having different forms or types than the ones considered in this study, Pareto front curves do not necessarily dominate each other, that is, some could be superior to others for a specific range of stability robustness while being inferior to the others for another range. In general, Pareto methods can precisely characterize which structure is superior for which range of objectives. Consequently, it can be emphasized that the Pareto front approach provides an objective comparison tool for interaction controllers of different forms and types, since it allows the designer to study, inspect, and compare all optimal controllers and trade-offs simultaneously.

We demonstrated the practical implementation of our approach for the design of
integer and fractional order admittance controllers. A fair comparison between
FOAC and IOAC is also presented, thanks to the Pareto front curves. Analysing the Pareto
front curves suggested that the curve for IOAC is inferior to that of FOAC.
Hence, FOAC allowed a better compromise between stability robustness and
transparency. Moreover, the results of our computational analysis showed that,
for the same robustness margin, $\alpha = 0.4$ results in the lowest parasitic
impedance allowing the highest transparency among the integration orders
considered in this study.

\begin{figure}[t]
    \centering
    \resizebox{0.90\columnwidth}{!}{\rotatebox{0}{\includegraphics[clip=true]{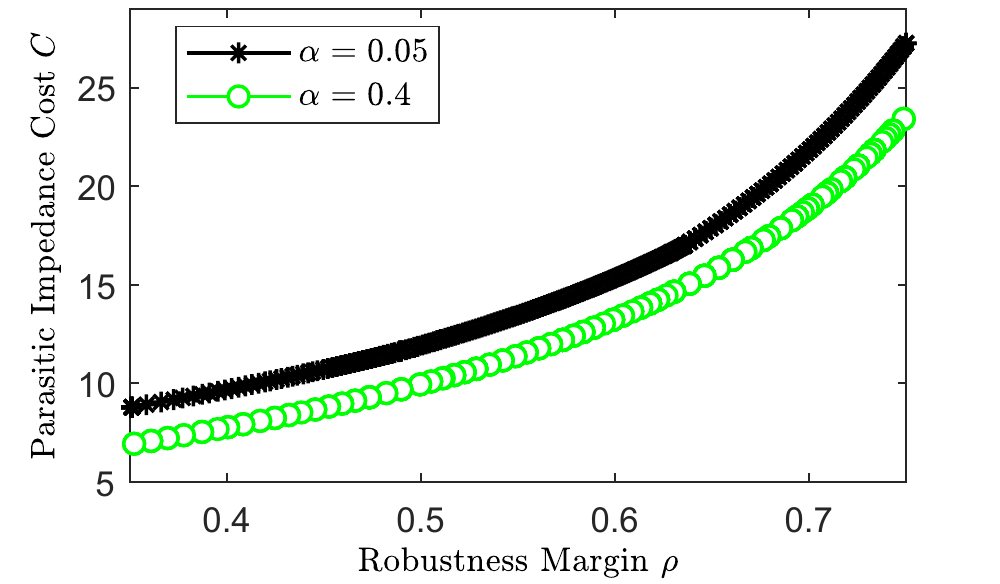}}}
    \vspace{-\baselineskip}
    \caption[]{Pareto front curves for FOAC with $\alpha=$0.4 and $\alpha=$0.05.\label{Fig:paretoDisc}}
    \vspace{-\baselineskip}
\end{figure}

To further study the effect of integration order, we computed the Pareto front curve of FOAC with $\alpha=0.05$. This Pareto front curve is presented in Figure~\ref{Fig:paretoDisc} together with the Pareto front curve for FOAC with $\alpha=0.4$. It can be observed from Figure~\ref{Fig:paretoDisc} that Pareto front curve of FOAC with $\alpha = 0.4$ is also superior than that of FOAC with $\alpha = 0.05$. These results indicate that the effect of integration order is not trivial and there exists an optimal integration order in terms of the objective functions considered. While $\alpha = 0.4$ provides the best value among the integration orders considered, further analysis is required to compute the optimal order.

We also verified the computational design by an
experimental pHRI task that involved contact interactions with a nonlinear environment formed by a two-stage spring. This one-dimensional scenario was chosen for the following reasons: i) The computational design of the controllers had been conducted for a single-input single-output (SISO) LTI system; hence, the experiments should match this design for a rigorous verification. ii) It is easier to conduct, quantify and present human subject experiments with relatively simple tasks, as more complex tasks are likely to introduce confounding effects. iii) This scenario captures the sudden changes in environment parameters, emulating a worst-case interaction as the total stiffness of the human-robot-environment system has been shown to have the most dominant effect on the coupled stability. iv) This scenario represents a simplified version of many practical tasks involving contact interactions with various environments, such as needle insertion or drilling.

Needle insertion is a commonly used task in many medical procedures, including biopsies, brachytherapy, and epidural anesthesia. During a needle insertion, the needle is advanced into the soft tissue by maintaining a certain direction as the deviation of the needle in other directions can break the needle with undesired consequences. Along these lines, a needle insertion is mostly a one-dimensional task. During a needle insertion task, the operator advances the needle into the tissue by appreciating the magnitude of the interaction forces. A two-stage spring system emulates the essence of this interaction as follows: The insertion of a needle into soft tissue starts with the deformation of the soft tissue (displaying a relatively stiff behaviour) and this is captured by the compression of Spring~1 in the experimental scenario. This deformation continues until rupture occurs, where a mixed stage of penetration and deformation starts. At this stage, as the needle continues its movement through the soft tissue, the forces tend to increase (displaying a different stiffness than the previous layer) until the needle reaches to another layer and this is captured by the activation of Spring~2 after a certain depth of compression.

To summarize, the experimental scenario in this study is considered as a representative simplification of a practical one-dimensional task. This choice enables experimental validation of the computational design, rigorous comparison of performance of the controllers considered and increased the digestibility of the procedure and the results. While it may be possible to test the proposed approach on other scenarios that require execution in multi-dimensions, such an extension will require extra effort to extend the proposed computational methodology. Such extensions are not considered as part of this study and will be left as a future research direction. Similarly, nonlinear or adaptive controllers are not considered in order to utilize powerful analysis tools available for LTI systems. We plan to investigate adaptive interaction controllers as part of our future work as suggested in \mbox{\cite{Ficuciello2015}}.

\vspace{-.5\baselineskip}

\section*{Acknowledgment}

The authors would like to thank to Prof. A. Kucukyilmaz for her valuable
comments. Moreover, Y.A. thanks to A. Aytekin, O. Caldiran, E.S. Emgin, G.
Serhat, and C. Yesil for fruitful discussions during the study. The Scientific
and Technological Research Council of Turkey (TUBITAK) supported this work under
contract EEEAG-117E645.

\vspace{-1\baselineskip}

\ifCLASSOPTIONcaptionsoff
    \newpage
\fi



%

\bibliographystyle{IEEEtran}
\bibliography{references}

%

\vspace{-1.5\baselineskip}
\begin{IEEEbiography}[{\includegraphics[width=1in,height=1.25in,clip,keepaspectratio]{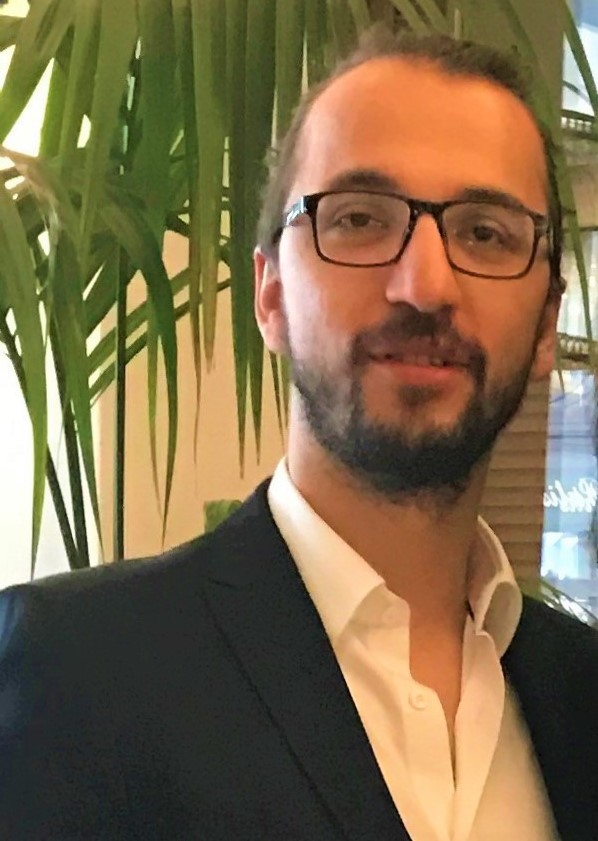}}]{Yusuf Aydin}
    Yusuf Aydin is currently a postdoctoral research fellow at Robotics and Mechatronics Laboratory at Koc University, Istanbul. He received his BSc dual degree in mechanical engineering and electrical and electronics engineering, MSc degree in mechanical engineering, and then, PhD degree in mechanical engineering from Koc University, Istanbul, in 2011, 2013, and 2019, respectively. He was awarded the prestigious TUBITAK BIDEB fellowship for his graduate studies. His research interests include physical human-robot interaction, haptics, robotics, control, optimization, and mechatronics.
\end{IEEEbiography}
\vspace{-1.5\baselineskip}
\begin{IEEEbiography}[{\includegraphics[width=1in,height=1.25in,clip,keepaspectratio]{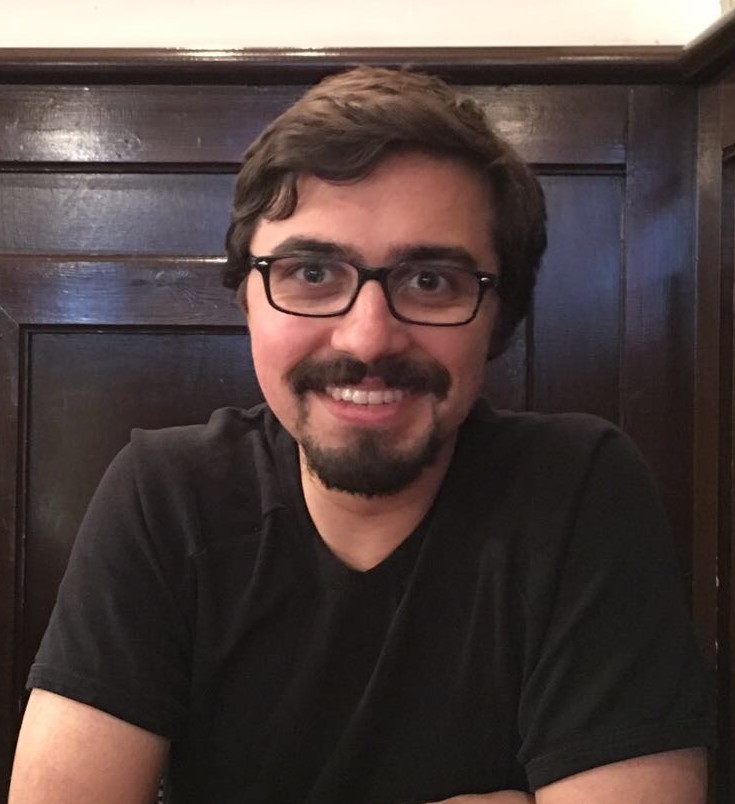}}]
    {Ozan Tokatli}
    Ozan Tokatli received his BSc, MSc and PhD degrees in Mechatronics Engineering from Sabanci University in 2008, 2010 and 2015, respectively. He worked as a post doctoral research associate in Haptics Laboratory at University of Reading. He is currently a research engineer in RACE, UKAEA which is a partner of the RAIN Hub. During his PhD, he worked on haptic systems with fractional order controllers and during his post-doctoral research, he investigated the use of haptic interfaces for classroom use to enhance the learning. His current research interest is physical human robot interaction in hazardous environments. His research interests extend to design of haptic interfaces and control of robots for haptics and pHRI.
\end{IEEEbiography}

\vspace{-1.5\baselineskip}
\begin{IEEEbiography}[{\includegraphics[width=1in,height=1.25in,clip,keepaspectratio]{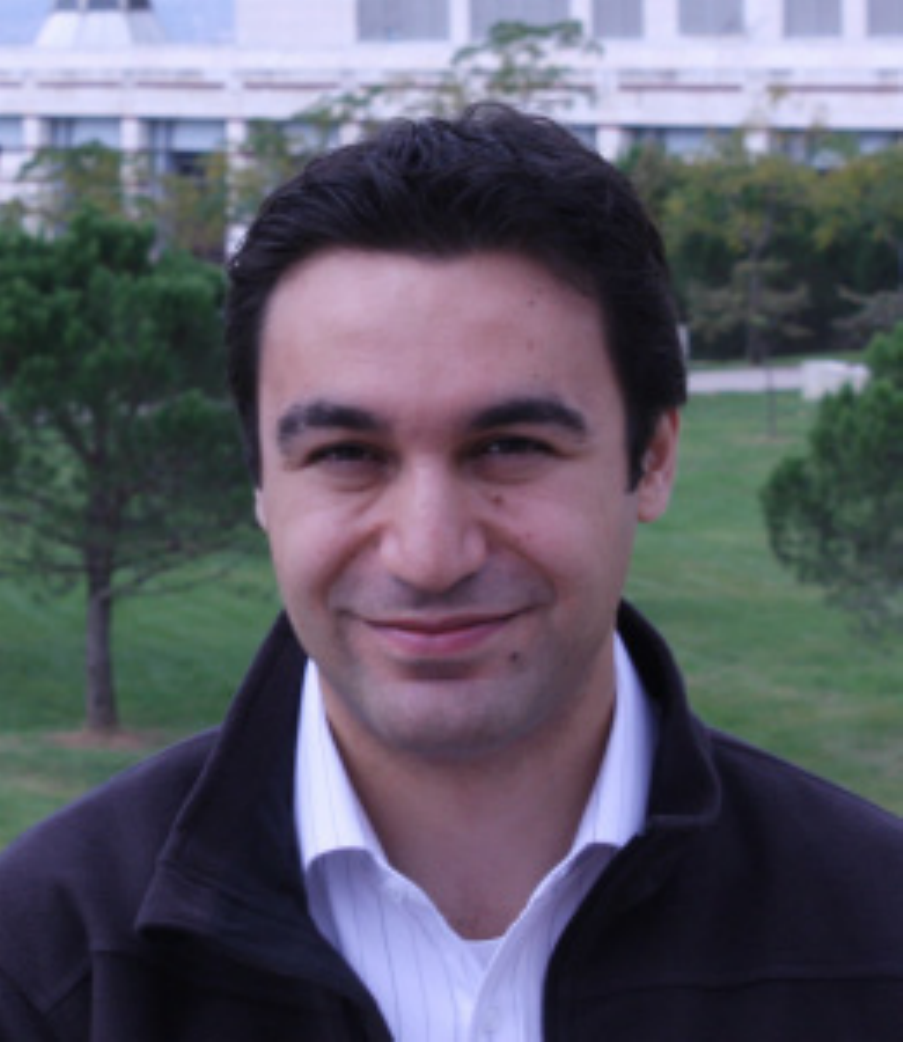}}]
    {Volkan Patoglu} Volkan Patoglu received his Ph.D. degree in
Mechanical Engineering from the University of Michigan, Ann Arbor in
2005. He worked as a post doctoral research associate at Rice University.
Currently, he is a professor at Sabanci University. His
research is in the area of physical human-machine interaction, in
particular, design and control of force feedback robotic systems
with applications to rehabilitation. His research
extends to cognitive robotics. Dr. Patoglu has served as an associate editor for  IEEE Transactions on Haptics (2013--2017) and is  an associate editor for  IEEE Transactions on Neural Systems and Rehabilitation Engineering and IEEE Robotics and Automation Letters.
\end{IEEEbiography}
\vspace{-1.5\baselineskip}
\begin{IEEEbiography}[{\includegraphics[width=1in,height=1.25in,clip,keepaspectratio]{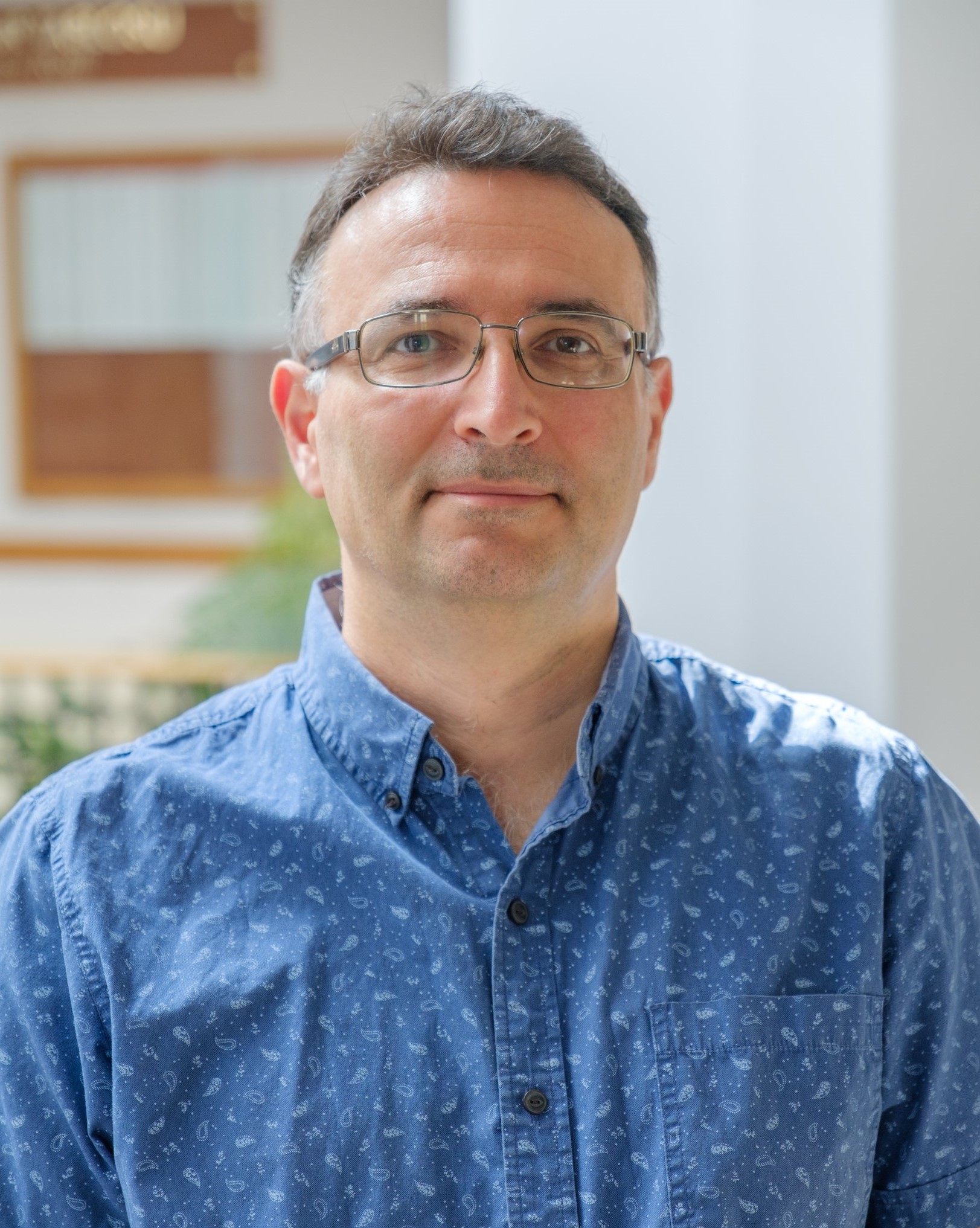}}]{Cagatay Basdogan}
    Cagatay Basdogan received the PhD degree in mechanical engineering from Southern Methodist University in 1994. He is a faculty member in the mechanical engineering and computational sciences and engineering programs of Koc University, Istanbul, Turkey. He is also the director of the Robotics and Mechatronics Laboratory at Koc University. Before joining Koc University, he worked at NASA-JPL/Caltech, MIT, and Northwestern University Research Park. His research interests include haptic interfaces, robotics, mechatronics, biomechanics, medical simulation, computer graphics, and multi-modal virtual environments. He is currently the associate editor in chief of IEEE Transactions on Haptics and serves in the editorial boards of IEEE Transactions on Mechatronics, Presence: Teleoperators and Virtual Environments, and Computer Animation and Virtual Worlds journals.
\end{IEEEbiography}


\vfill


\end{document}